\documentclass[10pt]{article}
\usepackage{fancyhdr}
\usepackage{extramarks}
\usepackage{amsmath}
\usepackage{amsthm}
\usepackage{amsfonts}
\usepackage{siunitx}
\usepackage{tikz}
\usepackage{algorithm}
\usepackage{algpseudocode}
\usepackage{multirow}
\usepackage{booktabs}
\usepackage{palatino}
\usepackage{graphicx}
\usepackage{subfigure}
\usepackage[colorlinks,linkcolor=black,anchorcolor=black,citecolor=black,urlcolor=blue]{hyperref}
\usepackage{amsmath,bm}
\usepackage{booktabs}
\usepackage{mathtools}
\usepackage{amssymb}
\usepackage{caption}
\usepackage{capt-of}
\usepackage{mciteplus}
\usepackage{cite}
\usepackage{mathrsfs}
\usepackage{parskip}
\usepackage{soul}
\usepackage{textcomp}
\usepackage[colaction]{multicol}
\usepackage[switch]{lineno}
\usepackage{lipsum}
\usepackage{etoolbox}
\usepackage{longtable}
\usepackage{array}
\usepackage{tablefootnote}
\usepackage{ragged2e}
\usepackage{lscape}
\newcolumntype{C}[1]{>{\centering\arraybackslash}p{#1}}
\usetikzlibrary{automata,positioning}
\usetikzlibrary{automata,positioning}

\usepackage{tikz,xcolor,hyperref}
\definecolor{lime}{HTML}{A6CE39}
\DeclareRobustCommand{\orcidicon}{%
	\begin{tikzpicture}
	\draw[lime, fill=lime] (0,0) 
	circle [radius=0.16] 
	node[white] {{\fontfamily{qag}\selectfont \tiny ID}};
	\draw[white, fill=white] (-0.0625,0.095) 
	circle [radius=0.007];
	\end{tikzpicture}
	\hspace{-2mm}
}
\foreach \x in {A, ..., Z}{%
	\expandafter\xdef\csname orcid\x\endcsname{\noexpand\href{https://orcid.org/\csname orcidauthor\x\endcsname}{\noexpand\orcidicon}}
}
\begin{document}

\title{Integrating Transformer and Autoencoder Techniques with Spectral Graph Algorithms for the Prediction of Scarcely Labeled Molecular Data}

%%%%%%%%%%%%%%%%%%%%%%%%%%%% Authors %%%%%%%%%%%%%%%%%%%%%%%%%%%%
\author{Nicole Hayes$^1$, 
Ekaterina Merkurjev$^{1,2}$\footnote{Corresponding author,
	Email:  merkurje@msu.edu} ~ and 
 Guo-Wei Wei$^{1,3,4}$\\
$^1$ Department of Mathematics, \\
Michigan State University, MI 48824, USA.\\
$^2$ Department of Computational Mathematics, Science and Engineering\\
Michigan State University, MI 48824, USA.\\
$^3$ Department of Electrical and Computer Engineering,\\
Michigan State University, MI 48824, USA. \\
$^4$ Department of Biochemistry and Molecular Biology,\\
Michigan State University, MI 48824, USA. \\
}
\date{}
\maketitle

%\title{Integrating Transformer and Autoencoder Techniques with Spectral Graph Algorithms for the Prediction of Scarcely Labeled Molecular Data}
%
%\author{Nicole Hayes \and     Ekaterina Merkurjev  \and Guo-Wei Wei }
%
\abstract{ 
In molecular and biological sciences, experiments are expensive, time-consuming, and often subject to ethical constraints. Consequently, one often faces the challenging task of predicting desirable properties from small data sets or scarcely-labeled data sets. Although transfer learning can be advantageous, it requires the existence of a related large data set. This work introduces three graph-based models incorporating Merriman-Bence-Osher (MBO) techniques to tackle this challenge. Specifically, graph-based modifications of the MBO scheme are integrated with state-of-the-art techniques, including a home-made transformer and an autoencoder, in order to deal with scarcely-labeled data sets. In addition, a consensus technique is detailed. The proposed models are validated using five benchmark data sets. We also provide a thorough comparison to other competing methods, such as support vector machines, random forests, and gradient boosting decision trees, which are known for their good performance on small data sets. The performances of various methods are analyzed using residue-similarity (R-S) scores and R-S indices. Extensive computational experiments and theoretical analysis show that the new models perform very well even when as little as 1\% of the data set is used as labeled data. 
}


%\tableofcontents

\section{Introduction}

In recent years, machine learning has significantly impacted many fields, including science, technology, and medicine \cite{jordan_ml, kotsiantis, schwab}. For instance, machine learning and deep learning methods have been applied to drug discovery, which has traditionally been a long and expensive process \cite{gnc}. In particular, drug discovery requires the knowledge of an assortment of molecular properties, such as toxicity, permeability, binding affinity, solubility, etc. Determining these properties experimentally is costly and time-consuming and may involve ethical concerns, so \textit{in silico} (i.e., computational) methods for predicting such molecular properties have become popular \cite{nature, gnc}. This work develops new computational techniques for making accurate predictions of many kinds of molecular properties.  

Overall, a significant challenge of machine learning algorithms, and in particular deep learning methods, is their reliance on a sufficient amount of labeled data to make accurate predictions about unlabeled data \cite{merkurjev2021multiscale, nature}. Because of the complexity, cost, and time required to obtain molecular information experimentally, it is often difficult to obtain large labeled molecular data sets \cite{jiang2020, hudson2000, saha2016, shaikhina2015, shaikhina2017}; this signifies a need for algorithms that can accurately predict molecular properties with limited labeled data \cite{hundreds, merkurjev2021multiscale}.  

One technique which has demonstrated predictive success for small scientific data sets is transfer learning, which aims to improve the performance of machine learning methods on a target domain by transferring the knowledge contained in a different but related domain; in this case, it is required that the two tasks have a relationship between each other. For example, the information obtained while learning to identify passenger cars could be used when trying to recognize pickup trucks or other large vehicles. Overall, transfer learning techniques can be viewed from the model viewpoint \cite{zhuang}. Transfer learning techniques that incorporate model-level regularizers in the objective function of the learner include the domain adaptation machine \cite{duan,duan2}, the consensus regularizer \cite{luo,zhuang2009}, and the domain-dependent regularizer \cite{ev,kato2007}. Other transfer learning techniques, such as parameter sharing \cite{zhuang2011,long2012} and parameter restriction \cite{tommasi}, focus on the parameters of the models. In addition, many scientists use various deep learning procedures to build their transfer learning models. Some examples of transfer learning approaches using deep learning techniques include transfer learning with deep autoencoders \cite{zhuang2017}, transfer learning motivated by deep residual learning \cite{long}, a selective adversarial network for partial transfer learning \cite{cao2018}, domain-adversarial neural network \cite{ganin2016}, and domain adversarial networks for the multisource transfer learning problem \cite{zhao2018}. 

Jiang et al. \cite{jiang2020} improved on the performance of existing transfer learning and deep learning methods for small data sets by introducing a boosting tree-assisted multitask deep learning (BTAMDL) framework, which consists of multitask deep transfer learning and gradient boosting decision trees (GBDT) in sequence. GBDT typically performs well on small data sets, while deep learning performs better on larger data sets, and in fact can perform significantly worse than traditional machine learning methods on small data sets \cite{feng2019}. The BTAMDL model thus pairs small scientific data sets with related large data sets and utilizes multitask deep transfer learning to simultaneously learn the target and source tasks (involving the small and large data sets, respectively) and transfer knowledge from the source to the target task. Unlike traditional multitask learning, which may attempt to benefit both related tasks, the BTAMDL framework only focuses on improving the performance on the small data set. Although the BTAMDL framework outperformed state-of-the-art methods on selected small data sets, its applications are limited to cases in which there exists a large data set that is sufficiently similar to the target small data set. Also, the performance of BTAMDL relies on the quality of the large datasets, which may be difficult to ensure in practice \cite{jiang2020}.  

In addition to transfer learning, active learning has been proposed for data sets with a small amount of labeled elements. In particular, active learning is a subset of machine learning which designs techniques that are able to interactively query a user to label some data elements. Specifically, in active learning, a method selects a subset of the most useful unlabeled data elements, which the human annotator then labels. The overall goal of active learning is to maximize a model’s performance gain while using the fewest labeled data samples possible.  Active learning techniques can be divided into membership query synthesis \cite{angluin,king2004}, stream-based selective sampling \cite{dagan1995,krishna}, and pool-based active learning from application scenarios \cite{lewis1994, settles2009}. Since deep learning methods often require a lot of labeled data, many deep active learning works have been developed. Some deep active learning techniques include deep Bayesian active learning \cite{gal2017}, deep active learning for link prediction in knowledge graphs \cite{ostapuk2019}, farthest first active learning \cite{geifman2017}, active learning with incremental neural architecture search \cite{geifman2019}, and deep reinforcement active learning \cite{liu2019}. 

However, the proposed methods of this work utilize a self-supervised learning approach for cases of low label rates. While traditional supervised machine learning methods require a labeled training data set, from which the algorithm attempts to learn the underlying pattern or structure of the data, self-supervised learning (SSL) uses unlabeled data, making it useful in cases where labeled data is difficult to acquire, such as with scientific data sets \cite{cluster-de}. SSL selectively masks portions of unlabeled data, allowing the model to be trained in a supervised manner. One example of a self-supervised technique is one by Chen et al. \cite{hundreds}, which involves a self-supervised learning framework to pretrain models from hundreds of millions of unlabeled molecules from multiple publicly available chemical databases. Their proposed bidirectional encoder transformer model, which generates molecular fingerprints that can subsequently be passed to machine learning algorithms for prediction, demonstrated state-of-the-art performance on regression and classification tasks involving molecular data sets. Although their SSL platform was pretrained on unlabeled data, the downstream tasks require labeled data; in this particular study, Chen et al. used 90\% of each molecular data set as their training set. One of the present aims is to examine the applicability of this transformer model on much lower amounts of labeled data. In particular, the proposed methods incorporate both autoencoder and transformer methods, to be discussed shortly in more detail, which do not require labeled data to generate latent-space vectors and can be used as fingerprints for molecular prediction. Furthermore, the proposed techniques incorporate variants of a graph-based Merriman-Bence-Osher (MBO) technique \cite{mbo, merkurjev2021multiscale, kostic}, which has been shown to perform successfully on data sets with low amounts of labeled data, obtaining accurate results with at most 1\% to 5\% of data serving as the training set \cite{mbo, merkurjev2021multiscale}.   

In addition, the proposed techniques incorporate molecular fingerprints for multi-task learning. In fact, methods using molecular fingerprints based on three-dimensional (3D) structures, which capture molecular spatial and structural information, have recently been successful in molecular classification and prediction problems \cite{nature, hundreds, 2D-valuable}. However, 3D molecular data sets are complex and suffer from high dimensionality. Furthermore, methods using 3D fingerprints have limited applicability due to their reliance on the availability of reliable 3D molecular data, which can be difficult to obtain \cite{nature, hundreds}. Two-dimensional (2D) molecular fingerprints are easier to obtain and less computationally complex because they do not depend on 3D structural information, such as stereochemistry of molecules \cite{nature, hundreds, 2D-valuable}. Moreover, Gao et al. \cite{2D-valuable} observed that, while this lack of complexity in 2D fingerprints makes them less effective than 3D fingerprints in predictions involving macromolecules like proteins, 2D-fingerprint-based models perform as well as 3D-fingerprint-based models in predicting toxicity, solubility, partition coefficient, and protein–ligand binding affinity (for models based only on ligand structure). 

Therefore, the proposed algorithms incorporate 2D molecular fingerprints. Overall, there are numerous methods for generating fingerprints from 2D molecular data sets, including  four main categories: substructure key-based fingerprints, topological or path-based fingerprints, circular fingerprints, and pharmacophore fingerprints \cite{2D-valuable, nature}. Another method uses an autoencoder \cite{gao2021proteome}, which typically consists of two neural networks: an encoder and a decoder. The encoder converts an input, such as 2D molecular data, into a fixed-size latent-space representation. The decoder aims to reconstruct the latent-space representation to recover the input. This process is unsupervised, meaning it needs no labels for the data. Moreover, the latent-space representation generated by the encoder can be extracted as a fingerprint for other tasks. Similarly, Chen et al. \cite{hundreds} proposed a method for generating fingerprints using a transformer, which is based on the attention mechanism, for  SSL  on unlabeled data. Their usage of SSL eliminated the need for a complete autoencoder with the encoder-decoder framework, allowing them to encode molecular data using only an encoder, specifically a bidirectional encoder transformer. The transformer is more easily parallelizable and achieves faster training than recurrent neural network (RNN)-based autoencoder models. The method in \cite{hundreds} achieved very good results in both molecular classification and regression problems. 
 
Overall, in this work, we propose three  new machine learning models: an autoencoder coupled with an MBO scheme (AE-MBO), a bidirectional encoder transformer coupled with an MBO scheme (BT-MBO), and an extended-connectivity fingerprint (ECFP) algorithm coupled with an MBO scheme (ECFP-MBO). We also detail a consensus model (Consensus-MBO). We validate these models on five benchmark classification data sets. In particular, we evaluate our models' performance on various amounts of labeled data, including as little as 1\% labeled data. 
Our AE-MBO model combines the autoencoder model developed earlier \cite{gao2021proteome} and the graph-based MBO method  \cite{mbo}. The encoder from the autoencoder converts 2D molecular data in simplified molecular input line entry specification (SMILES) \cite{smiles} format to a latent-space representation. These latent-space vectors serve as autoencoder-based molecular fingerprints (AE-FPs), which are then used as features in the MBO method. Similarly, our BT-MBO model integrates the bidirectional encoder transformer developed earlier  \cite{hundreds} and the graph-based MBO method. The transformer converts SMILES input to latent-space vectors called bidirectional transformer-based fingerprints (BT-FPs), which are subsequently passed to the MBO algorithm for the classification task. Our ECFP-MBO method utilizes the circular fingerprint generator from the RDKit library \cite{rdkit}, which produces ECFPs that serve as an input to our MBO method. Additional details of the autoencoder, transformer, and ECFP models are discussed in Section \ref{Methods}, and detailed results of the methods on the five classification data sets are given in Section \ref{Results}.  

The main advantage of our proposed models is their ability to perform accurately even in cases of a small amount of labeled elements, a significant feat due to scarcity of labeled data for many applications.

\section{Methods} \label{Methods}

In this section, we discuss the details of the proposed AE-MBO (autoencoder with MBO), BT-MBO (bidirectional transformer with MBO), ECFP-MBO (extended-connectivity fingerprints with MBO), and Consensus-MBO methods used in our work. 

\subsection{Data Preprocessing}

Our proposed methods require data consisting of SMILES strings of molecular compounds as well as a label corresponding to each compound. In the present work, only binary classification problems are considered, but the process for the multiclass case is similar. We test our models on five data sets, summarized in Table \ref{tab:datasets}, each of which contains SMILES strings and labels in addition to other molecular data. Further details of the contents of the five data sets are discussed in Section \ref{Results}. Thus, we first accordingly extract only the necessary information from each data set: a vector of SMILES strings and a corresponding vector of binary labels. Upon obtaining these vectors for each data set, we then construct AE-FPs as discussed in Section \ref{Autoencoder}, BT-FPs as discussed in Section \ref{Transformer}, and ECFPs as discussed in Section \ref{ecfps}, which are then passed as features for our MBO algorithm as discussed in Section \ref{MBO}.

\subsection{MBO Scheme} \label{MBO}

The MBO scheme utilized in the present work is based on the techniques outlined in the literature \cite{mbo, merkurjev2021multiscale}, which adapt the original MBO algorithm \cite{mbo1994} to the important tasks of data classification and segmentation. Our MBO scheme relies on a similarity graph-based framework represented by a graph $G = (V, E)$, where $V$ is the set of vertices representing the data set elements, and $E$ is the set of edges connecting some pairs of vertices. Moreover, we consider a weight function $w: V\times V \to \mathbb{R}$, which measures the degree of similarity between pairs of vertices; the value $w(i,j)$ represents the degree of similarity between vertices $i$ and $j$. A large value of $w(i,j)$ indicates a high degree of similarity between vertices $i$ and $j$, while a small value of $w(i,j)$ indicates that the vertices are dissimilar. Overall, the choice of weight function depends on the data set at hand, but one commonly used weight function is the Gaussian function:
\begin{equation*}
    w(i,j) = \text{exp} \left( - \frac{d^2(i,j)}{\sigma^2} \right),
\end{equation*}
where $d(i,j)$ represents the distance between data elements represented by vertices $i$ and $j$, and $\sigma>0$ \cite{mbo, merkurjev2021multiscale}.\\

We can define a weight matrix $\mathbf{W}$ by $W_{ij} = w(i,j)$ and let the degree of a vertex $i \in V$ be $d_i = \sum_{j \in V} w(i,j)$. If we define $\mathbf{D}$ as the diagonal matrix with diagonal elements $\{d_i\}$, then the graph Laplacian is defined as
\begin{equation}
\label{laplacian}
    \mathbf{L} = \mathbf{D} - \mathbf{W}.
\end{equation}

The goal of the MBO algorithm for our data classification problem is to generate an optimal label matrix $\mathbf{U} = (\mathbf{u}_1, \dots , \mathbf{u}_N)^T$, where $\mathbf{u}_i \in \mathbb{R}^2$ represents the probability distribution over the classes for the data element represented by vertex $i$. Each vector $\mathbf{u}_i$ is an element of the Gibbs simplex with $m$ vertices, where $m$ is the number of classes:
\begin{equation}
\label{gibbs}
    \Sigma^m := \left\{(z_1, ..., z_m) \in [0,1]^m | \sum_{k=1}^m z_k = 1\right\}
\end{equation}

The vertices of this simplex are given by the unit vectors $\{\mathbf{e}_i\}$, with only one nonzero entry of $1$, which represent nodes belonging exclusively to each of the  classes.  

In general, many graph-based optimization-based data classification problems aim to minimize the energy $E(\mathbf{U}) = R(\mathbf{U}) + \text{Fid}(\mathbf{U})$, where $\mathbf{U}$ is the data classification matrix, with each row representing a probability distribution of a data element over all classes, $R(\mathbf{U})$ is a regularization term that contains the graph weights, and $\text{Fid}(\mathbf{U})$ is a fidelity term that incorporates labeled points or other information \cite{mbo, merkurjev2021multiscale}.  

One term that can be used for $\text{Fid}(\mathbf{U})$ is an $L_2$ fit to all elements with known class, i.e., all labeled elements. For the regularization term $R(\mathbf{U})$, one can consider a variant of the Ginzburg-Landau functional \cite{mbo, merkurjev2021multiscale}, which is related to the total variation functional often used in image processing applications. To minimize the energy $E$ in the $L_2$ sense in the continuous case, one can apply the gradient descent procedure and obtain a modified Allen-Cahn equation containing a forcing term. Applying a time-splitting scheme to this equation yields an algorithm that alternates between diffusion and thresholding steps, where the diffusion step involves the heat equation with a forcing term. In a graph-based framework, the Laplace operator in the heat equation can be replaced by a version of the graph Laplacian (\ref{laplacian}). For both the binary and multiclass case, the thresholding step can be replaced by displacement toward the closest vertex in the Gibbs simplex (\ref{gibbs}), as further derived in the literature \cite{mbo, merkurjev2021multiscale}.  \\ \\

%\begin{figure*}[h!]
\begin{algorithm}[h!]
\caption{MBO Algorithm (based on techniques in \cite{merkurjev_aml, merkurjev2021multiscale})}\label{alg-mbo}
%\rule{18cm}{0.2mm}
\begin{algorithmic}[H]
\vspace{0.075cm}
\Require  Labeled data $\mathscr{L} = \{(\mathbf{x}_i, y_i)\}_{i=1}^N$, where $y_i$ is the label of the compound whose fingerprint is $\mathbf{x}_i$, $N$ (\# of data set elements), $N_n$ (\# of nearest neighbors), $N_e$ (\# of eigenvectors to be computed), $C$, $dt > 0$, $N_t$ (maximum \# of iterations), $N_p$ (\# of labeled points or \# of points in training set), $N_l$ (\# of labeled sets), $N_s$ (\# of times to apply diffusion operator in MBO method), $m$ (\# of classes), $\Gamma$ (vector with $1$ for labeled nodes and $0$ otherwise).
% \vspace{0.1cm}
\Ensure  $\mathbf{u}^{\text{end}}$, where each row of $\mathbf{u}^{\text{end}}$ is a probability distribution of data point $\mathbf{x}_i$ over the two classes.
\vspace{0.1cm}
\State 1: Construct graph with $N_n$ nearest neighbors or go straight to Step 3 using the Nystr\"{o}m technique \cite{nystrom1,nystrom2,nystrom3} (useful for very large data sets).
\vspace{0.1cm}
\State 2: Compute the graph Laplacian $\mathbf{L}$.
\vspace{0.1cm}
\State 3: Compute (or approximate using the Nystr\"{o}m method) the first $N_e$ eigenvalues $\Lambda$ and eigenvectors $\Phi$ of the graph Laplacian.
\vspace{0.1cm}
\State 4: Complete the following steps:
\vspace{0.3cm}
\For{$i = 1 \to N_l$}
\vspace{0.1cm}
\State Randomly generate $N \times 1$ vector $\Gamma$, where $\Gamma_j = 1$ for $N_p$ labeled points and $\Gamma_j = 0$ otherwise.
\vspace{0.1cm}
\State $\mathbf{U}_{i k}^{~0} \leftarrow rand((0,1))$, except for labeled nodes, where $\mathbf{u}_i^0 \leftarrow \boldsymbol{e}_k$ for labeled nodes, where $k$ is the true class.
\vspace{0.1cm}
\State $\mathbf{u}_i^0 \leftarrow projectToSimplex(\mathbf{u}_i^0)$, where $\mathbf{u}_i$ is $i^{th}$ row of $\mathbf{U}^0$.
\vspace{0.1cm}
%\State (Main MBO algorithm (expand), output is probability distribution over the 2 classes)
\vspace{0.1cm}
\State $\mathbf{U} \leftarrow \mathbf{U}^0$
\vspace{0.1cm}
\State $\mathbf{A} \leftarrow \Phi'\mathbf{U}$,  \hspace{0.2cm} $\mathbf{B} \leftarrow \mathbf{0}$, \hspace{0.2cm}  $\mathbf{E} \leftarrow 1 + (dt/N_s)\Lambda$
\vspace{0.2cm}
\For{$i = 1 \to N_t$}
\vspace{0.1cm}
\For{$k = 1 \to N_s$}
\vspace{0.1cm}
\For{$j = 1 \to m$}
\vspace{0.1cm}
\State $\mathbf{A}_j \leftarrow (\mathbf{A}_j- (dt/N_s)\mathbf{B}_j)./\mathbf{E}$
\vspace{0.1cm}
\EndFor
\vspace{0.1cm}
\State $\mathbf{u}_i^{n+1} \leftarrow \Phi \mathbf{A}$
\vspace{0.1cm}
\State $\mathbf{B} \leftarrow \mathbf{C} (\Phi' (\Gamma \cdot (\mathbf{U} - \mathbf{U}^0))) $, with $\cdot$ indicating row-wise multiplication.
\vspace{0.1cm}
\EndFor
\vspace{0.1cm}
\State $\mathbf{u}_i^{n+1} \leftarrow projectToSimplex(\mathbf{u}_i^{n+1})$
\vspace{0.1cm}
\State $\mathbf{u}_i^{n+1} \leftarrow \boldsymbol{e}_k$, where $k$ is closest simplex vertex to $\mathbf{u}_i^{n+1}$ (i.e., column of $\mathbf{u}_i^{n+1}$ with highest probability)
\vspace{0.1cm}
\EndFor
\vspace{0.1cm}
\EndFor
\vspace{0.3cm}
\end{algorithmic}
%\rule{18cm}{0.2mm}
\end{algorithm}
%\end{figure*}

Accordingly, the MBO algorithm used in the present work can be summarized as follows. Given the input data (AE-FPs, BT-FPs, or ECFPs, along with labels), a graph with a specified number of nearest neighbors $N_n$ is constructed. The graph Laplacian and a specified number $N_e$ of eigenvalues and eigenvectors are computed from the graph. A subset of the input data is selected as the training set, or labeled  set, and the remainder is unlabeled for testing. An initial guess for $\mathbf{U}$ is randomly generated, where the rows corresponding to labeled points are given by their respective unit vectors $\{\mathbf{e}_i\}$ in (\ref{gibbs}). Then, each successive iterate of $\mathbf{U}$ is obtained via the following steps: 
\vspace{0.2cm}
\begin{enumerate}
    \item Diffusion: obtain $\mathbf{U}^{n+\frac{1}{2}}$ from heat equation with forcing term, applied $N_s$ times.
    \item Projection to simplex: obtain $\mathbf{U}^{n+1}$ by projecting each row of $\mathbf{U}^{n+\frac{1}{2}}$ onto (\ref{gibbs}).
    \item Displacement: Each row of the projected matrix from step 2 is replaced by its closest vertex in the simplex (\ref{gibbs}).
\end{enumerate}
\vspace{0.2cm}
This process is repeated for a set number of iterations $N_t$ to obtain a class prediction for each data point. 

We note that the scheme's efficiency is increased by using spectral techniques and a low-dimensional subspace spanned by only a small number of eigenfunctions. In particular, the goal is to write the matrix $\mathbf{U}$ as the product $\Phi \mathbf{A}$, where $\Phi$ is the matrix of $N_e$ eigenvectors of the graph Laplacian and $\mathbf{A}$ is a coefficient matrix to be determined. The forcing term in the diffusion step can also be written in terms of the $N_e$ eigenvectors of the graph Laplacian. This spectral technique allows the diffusion operator of the MBO method to be decomposed into faster matrix multiplications, while sacrificing little accuracy in practice.

In addition, the MBO procedure can be efficiently used for very large data sets using the Nystr\"{o}m technique \cite{nystrom1,nystrom2,nystrom3}. In particular, the Nystr\"{o}m procedure allows the approximation of the $N_e$ eigenvectors of the graph Laplacian without the need to calculate all the graph weights; in fact, only a very small portion of them need to be computed. 

The details of this MBO method are outlined in Algorithm \ref{alg-mbo}.

\subsection{Autoencoder} \label{Autoencoder}

We now outline the procedure for generating AE-FPs, which is summarized in Algorithm \ref{alg-auto}.

An autoencoder typically consists of two neural networks: an encoder and a decoder. The encoder aims to convert some input (in this case, in simplified molecular-input SMILES format) to a latent space vector of a specified length (in this case, 512). The decoder then attempts to recover the initial input from its latent space representation. Specifically, the decoder converts the latent space representation to the probability distribution of the target, which can in turn predict the target of interest (for instance, the SMILES string of a molecule). The autoencoder is trained to minimize the error of this prediction \cite{hundreds}. Because the encoder and decoder do not require data labels to construct the latent space representation of the input or the associated prediction, autoencoders are unsupervised, allowing for prediction of unlabeled data sets.

The intermediate latent space vector in an autoencoder corresponding to each input sequence can be thought of as encoding the "meaning" of that input sequence \cite{gnc}. In fact, the latent space vectors are often used as fingerprints for other tasks, such as the prediction of various molecular properties \cite{zupan}. In such cases, the decoder does not contribute to the final prediction and instead only aids in training the network \cite{hundreds}. In the present work, we take this approach, using the autoencoder network developed by {\color{black} Gao et al.} in an earlier work \cite{gao2021proteome}, in which the encoder and decoder are comprised of long short-term memory (LSTM) networks, to generate a fingerprint (AE-FP) corresponding to each input molecule. The fingerprints can then be used as features in any data classification algorithm to predict the label for each molecule. In {\color{black} this work}, we propose utilizing an MBO scheme, outlined in Section \ref{MBO}, to classify the molecules, resulting in an integrated AE-MBO model. We also compare the results of our proposed model with methods using the AE-FPs as features in other data classification algorithms in Section \ref{Results}. {\color{black} Before passing the AE-FPs to the MBO algorithm and the comparison models for classification, we scale the features to have zero mean and unit variance.}

%\begin{figure*}[h!]
\begin{algorithm}[h!]
\caption{AE-MBO Algorithm (incorporating similar techniques as
in \cite{gao2021proteome})}\label{alg-auto}
%\rule{18cm}{0.2mm}
\begin{algorithmic}[H]
\vspace{0.075cm}
\Require  Unlabeled data $\mathscr{U} = \{s_i\}_{i=1}^N$, where $s_i$ is the SMILES string of a molecular compound, pretrained LSTM model, a dictionary that assigns an integer value to each SMILES symbol.
\vspace{0.1cm}
\Ensure  Set of fingerprints $\mathscr{F}_{\text{AE}} = \{\mathbf{x}_i\}_{i=1}^N$.
\vspace{0.1cm}
\State 1: Load pretrained LSTM model.
\vspace{0.1cm}
\State 2: Extract hidden features from $\mathscr{U}$ using the pretrained model:
\vspace{0.1cm}
\State Initialize empty list of hidden features $\mathscr{F}_{\text{AE}}$ of length $N$.
\vspace{0.1cm}
\For{$i = 1 \to N$}
\vspace{0.1cm}
\State Generate a tensor $\mathbf{t}_i$ with dimension equal to the length of the string $s_i$, where each element of $\mathbf{t}_i$ is the integer dictionary value of the character in the corresponding position in $s_i$.
\vspace{0.1cm}
\State Encode $\mathbf{t}_i$ into a tensor $\mathbf{V}_i$ of shape $(1, 1, 512)$ using pretrained LSTM model.
\vspace{0.1cm}
\State Reshape $\mathbf{V}_i$ into a one-dimensional vector $\mathbf{x}_i$.
\vspace{0.1cm}
\State Append $\mathbf{x}_i$ to list of hidden features.
\vspace{0.1cm}
\EndFor
\vspace{0.1cm}
\State 3: Save set of fingerprints $\mathscr{F}_{\text{AE}} = \{\mathbf{x}_i\}_{i=1}^N$ from hidden features.
\vspace{0.1cm}
\State 4: {\color{black} Scale features $\mathscr{F}_{\text{AE}} = \{\mathbf{x}_i\}_{i=1}^N$ to have zero mean and unit variance.}
\vspace{0.1cm}
\State 5: Send {\color{black} scaled features} $\mathscr{F}_{\text{AE}} = \{\mathbf{x}_i\}_{i=1}^N$ to the MBO method in Algorithm \ref{alg-mbo}.
\vspace{0.1cm}
\end{algorithmic}
%\rule{18cm}{0.2mm}
\end{algorithm}
%\end{figure*}

\subsection{Transformer} \label{Transformer}

We now outline the process for generating BT-FPs, the details of which are summarized in Algorithm \ref{alg-bt}. 

Many models designed to process sequential data, as in language processing, utilize the encoder-decoder structure of the autoencoder described in Section \ref{Autoencoder} integrated with recurrent neural networks (RNNs) or convolutional neural networks (CNNs). It is advantageous to further connect the encoder and decoder via an attention mechanism, which informs the decoder which portions of the input are most relevant. Such models have demonstrated some of the best performance in sequential processing tasks \cite{vaswani}. Vaswani et al. \cite{vaswani} introduced a transformer model based solely on an attention mechanism, avoiding the reliance on RNNs and CNNs. Unlike RNNs, for instance, this proposed transformer model does not need to process sequential data in order, allowing for more parallelization and, consequently, faster training. 

The transformer model used in the present work was designed in an earlier work \cite{hundreds} and was constructed based solely on an attention mechanism for self-supervised learning (SSL), as in the literature \cite{vaswani}. The model in \cite{hundreds}, i.e., the self-supervised learning platform (SSLP), pretrains deep learning networks using over 700 million unlabeled SMILES strings. The platform constructs data-mask pairs by masking sections of each SMILES input, which allows the SSLP to train the networks in a supervised manner, despite the data itself being unlabeled. By using SSL, the model is able to encode SMILES input using only a bidirectional encoder transformer rather than a full encoder-decoder structure \cite{hundreds}.

Chen et al. \cite{hundreds} constructed three SSLP models trained on three different data sets: one model pretrained on ChEMBL \cite{chembl-source}, one model pretrained on the union of ChEMBL and PubMed \cite{pubchem-source}, and one model pretrained on the union of ChEMBL, PubMed, and ZINC \cite{zinc-source}. The sizes of these data sets range from over one million to over 700 million \cite{hundreds}. The results of the predictive tasks associated with these three models are given in Table \ref{tab:90 Comparison} and demonstrate that the model trained on the largest data set did not always have the best performance. In the present work, we use the SSLP model pretrained on ChEMBL only. In \cite{hundreds}, this model achieved the best predictive performance of the three pretrained models on three of the five benchmark data sets considered in this work. Additionally, Chen et al. fine-tuned their selected pretrained model for specific downstream tasks \cite{hundreds}. In the present work, we generated BT-FPs directly for each data set without fine-tuning the model pretrained on ChEMBL.

As in the AE-MBO model described in Section \ref{Autoencoder}, the proposed BT-MBO method converts SMILES input for a given molecular compound to a latent space vector, which is extracted as a fingerprint (BT-FP). The BT-FPs for all molecules in a data set are then {\color{black} compiled into a set of features for that data set. The features are scaled to have zero mean and unit variance and subsequently passed to} an MBO algorithm, which ultimately predicts the molecular labels. For comparison, we also passed these {\color{black} scaled} BT-FPs as features to other data classification algorithms, as outlined in Section \ref{Results}. 
\vspace{0.4cm}

%\begin{figure*}[h!]
\begin{algorithm}[H]
\caption{BT-MBO Algorithm (incorporating similar techniques as
in \cite{hundreds})}\label{alg-bt}
%\rule{18cm}{0.2mm}
\begin{algorithmic}[H]
\vspace{0.075cm}
\Require  Unlabeled data $\mathscr{U} = \{s_i\}_{i=1}^N$, where $s_i$ is the SMILES string of a molecular compound, a selected pretrained model (from model pretrained on ChEMBL, model pretrained on union of ChEMBL and PubMed, or model pretrained on the union of ChEMBL, PubMed, and ZINC), a dictionary that assigns an integer value to each SMILES symbol.
\vspace{0.1cm}
\Ensure  Set of fingerprints $\mathscr{F}_{\text{BT}} = \{\mathbf{x}_i\}_{i=1}^N$.
\vspace{0.1cm}
\State 1: Preprocessing: binarize input SMILES data.
\vspace{0.1cm}
\State 2: Load selected pretrained model.
\vspace{0.1cm}
\State 3: Extract hidden information from $\mathscr{U}$ using the pretrained model:
\vspace{0.1cm}
\State Initialize empty dictionary $D_\text{hidden}$ of length $N$ of hidden features.
\vspace{0.1cm}
\For{$i = 1 \to N$}
\vspace{0.1cm}
\State Generate a tensor $\mathbf{t}_i$ with dimension equal to the length of the string $s_i$, where each element of $\mathbf{t}_i$ is the integer dictionary value of the character in the corresponding position in $s_i$.
\vspace{0.1cm}
\State Pass $\mathbf{t}_i$ to the bidirectional encoder transformer from the pretrained model and extract the inner state from the last hidden layer $\mathbf{T}_i$, whose dimensions are $(l_i, 1, 512)$ (where $l_i$ is the length of $\mathbf{t}_i$).
\vspace{0.1cm}
\State Reshape $\mathbf{T}_i$ into a tensor $\mathbf{V}_i$ with dimensions $(l_i, 512)$.
\vspace{0.1cm}
\State Save $\mathbf{V}_i$ as $i$th element of $D_\text{hidden}$.
\vspace{0.1cm}
\EndFor
\vspace{0.1cm}
\State 4: Generate fingerprints from hidden information:
\vspace{0.1cm}
\State Initialize empty list of fingerprints $\mathscr{F}_{\text{BT}}$ of length $N$.
\vspace{0.1cm}
\For{$i = 1 \to N$}
\vspace{0.1cm}
\State Extract first row $\mathbf{x}_i$ from the $i$th element of $D_\text{hidden}$.
\vspace{0.1cm}
\State Save $\mathbf{x}_i$ as $i$th fingerprint in $\mathscr{F}_{\text{BT}}$.
\vspace{0.1cm}
\EndFor
\vspace{0.1cm}
\State 5: {\color{black} Scale features $\mathscr{F}_{\text{BT}} = \{\mathbf{x}_i\}_{i=1}^N$ to have zero mean and unit variance.}
\vspace{0.1cm}
\State 6: Send {\color{black} scaled} features $\mathscr{F}_{\text{BT}} = \{\mathbf{x}_i\}_{i=1}^N$ to the MBO method in Algorithm \ref{alg-mbo}.
\vspace{0.1cm}
\end{algorithmic}
%\rule{18cm}{0.2mm}
\end{algorithm}
%\end{figure*}

\subsection{Extended-Connectivity Fingerprints} \label{ecfps}

Extended-connectivity fingerprints (ECFPs) are a type of circular fingerprints used for molecular characterization. ECFPs aim to encode features that represent molecular activity. ECFPs have been shown to be useful in molecular prediction tasks involving drug activity \cite{circ-fp, hundreds}. The ECFP algorithm is a variation of the Morgan algorithm \cite{morgan}, which uses an iterative process to identify each atom in a molecule with a numerical label. The standard Morgan algorithm first records an initial atom identifier from numbering-invariant information about the atoms. Then, it generates new identifiers using the identifiers from the previous iteration, where new identifiers are independent of the original atom numbering in the molecule. This algorithm continues until the atom identifiers are as unique as possible, comprising a canonical numbering of the atoms, and the intermediate identifiers are subsequently discarded \cite{circ-fp}. 

Instead of proceeding until identifier uniqueness or near-uniqueness is achieved, the ECFP algorithm terminates after a set number of iterations. Also, the ECFP algorithm stores the initial atom identifiers and identifiers from intermediate steps rather than discarding them. In fact, the ECFP is defined by the set of initial, intermediate, and final atom identifiers. Because it does not aim for a unique, canonical numbering as the standard Morgan algorithm does, the ECFP algorithm is less computationally expensive and generates identifiers that are comparable between different molecules \cite{circ-fp}.

To generate ECFPs, we utilized functions from the RDKit library \cite{rdkit} that build circular fingerprints based on the Morgan algorithm. These RDKit functions require a radius parameter, which indicates the number of iterations (i.e., updates to the atom identifiers) the algorithm should perform. Each iteration considers a neighborhood of increasing radius centered at each atom to update that atom's identifiers. Accordingly, the RDKit radius parameter describes the radius of the largest neighborhood the algorithm considers for each atom. However, ECFPs are typically identified by the diameter, rather than the radius, used to generate them, where the diameter is then twice the number of iterations the algorithm performs \cite{circ-fp}. To remain consistent with ECFP naming conventions, we thus refer to all ECFPs generated in the present work using the diameter parameter. More information about the ECFP notation used in our models can be found in Section \ref{Parameters}.

{\color{black} As in the AE-MBO and BT-MBO methods, we pass the set of ECFPs for each data set as features to an MBO scheme, resulting in an integrated ECFP-MBO model. We also generated comparisons, reported in Section \ref{Results}, by passing the ECFPs as features to other data classification algorithms. Unlike in the AE-MBO and BT-MBO methods, we did not scale or otherwise normalize the ECFP features, since all entries in the ECFPs are either 0 or 1.} Further details of the ECFP-MBO model used in the present work are given in Algorithm \ref{alg-ecfp}.\\

%\begin{figure*}[h!]
\begin{algorithm}[H]
\caption{ECFP-MBO Algorithm (incorporating similar techniques as in \cite{circ-fp}, implemented using \cite{rdkit})}\label{alg-ecfp}
%\rule{18cm}{0.2mm}
\begin{algorithmic}[H]
\vspace{0.075cm}
\Require  Unlabeled data $\mathscr{U} = \{s_i\}_{i=1}^N$, where $s_i$ is the SMILES string of a molecular compound, diameter $d$, number of bits $N_\text{bits}$ (also called folding length).
\vspace{0.1cm}
\Ensure  Set of fingerprints $\mathscr{F}_{\text{ECFP}} = \{\mathbf{x}_i\}_{i=1}^N$.
\vspace{0.1cm}
\State 1: Initial assignment: assign each atom an integer identifier, ignoring hydrogen atoms and bonds to hydrogen atoms. Store identifiers in $\mathscr{F}_{\text{ECFP}}$.
\vspace{0.1cm}
\State 2: Iterative updating:
\For{$i = 1 \to \frac{d}{2}$}
\vspace{0.1cm}
\State Update each identifier to indicate (1) its neighbors' identifiers and (2) if it is a duplicate of other features.
\vspace{0.1cm}
\State Store identifiers in $\mathscr{F}_{\text{ECFP}}$.
\vspace{0.1cm}
\EndFor
\State 3: Duplicate identifier removal: reduce feature duplicates to a single feature.
\vspace{0.1cm}
\State 4: Send fingerprints $\mathscr{F}_{\text{ECFP}} = \{\mathbf{x}_i\}_{i=1}^N$ stored as bit vectors with length $N_\text{bits}$ to the MBO method in Algorithm \ref{alg-mbo}.
\vspace{0.1cm}
\end{algorithmic}
%\rule{18cm}{0.2mm}
\end{algorithm}
%\end{figure*}

\subsection{Consensus Model} \label{Consensus}

To attempt to achieve better performance among the proposed MBO-integrated methods, we introduce the Consensus-MBO method, a consensus model of our proposed methods. To generate a prediction using the consensus model for a given data set and percentage of labeled data, we first identify the two best-performing MBO-integrated models from among BT-MBO, AE-MBO, and ECFP-MBO (results displayed and discussed in Section \ref{Performance}). Then, instead of outputting the predicted labels, we output an $N \times 2$ matrix of probabilities from each of the two selected proposed methods, where the $(i,j)$th entry represents the probability that the $i$th data element is in class $j$. For each data element, we average these probabilities over the two methods and threshold the averaged probability to get the consensus prediction. In the binary classification task, this thresholding step amounts to assigning each data element to the class with higher averaged probability.

\section{Results and Discussion} \label{Results}

\subsection{Data Sets} \label{Data Sets}

We tested the proposed models on five benchmark classification data sets:
\begin{itemize}
    \item The Ames data set contains results of the bacterial reverse mutation assay (Ames test), which detects mutagenicity in vitro. The data set includes SMILES strings of 6,512 compounds paired with their binary labels (positive or negative mutagenicity) from the Ames test results \cite{ames_source}.
    \item The Bace data set consists of binding data for 1,513 inhibitors of human $\beta$-secretase 1 (BACE-1), namely the SMILES strings of the inhibitor compounds and their corresponding binary binding result labels \cite{bace_source}.
    \item The BBBP data set records results of a study on the blood-brain barrier permeability. The data set contains SMILES strings of 2,039 compounds and binary labels indicating each compound's penetration or non-penetration of the blood-brain barrier \cite{bbbp_source}.
    \item The Beet data set contains results of a study on toxicity of pesticides in honeybees. For each of the 254 compounds in the data set (given as SMILES strings), the study recorded an experimental value of the amount of active substance per bee, then classified compounds as positive if their experimental values were above a certain threshold and negative otherwise. The study used two different toxicity thresholds to generate two sets of binary labels. In the present work, we utilize the labels corresponding to the $100 \mu$/bee threshold \cite{beet_source} to be consistent with the results from Chen et al. \cite{hundreds}.
    \item The ClinTox data set lists records of drugs that have been approved by the FDA and drugs that have failed clinical trials due to toxicity. The data set consists of SMILES representations for 1,478 drugs along with two sets of binary labels representing their FDA approval status (approved or not approved) and their clinical trial results (toxicity or absence of toxicity). In the present work, we use labels corresponding to the clinical trial results \cite{clintox_source}.
\end{itemize}

Note that each data set sorts its elements into two classes. The details of these five data sets are summarized in Table \ref{tab:datasets}.\\

\begin{table}[h!]
\centering

 \begin{tabular}{c c c c}
 \hline
 Data Set & Description & \# Compounds & Source\\
 \hline
 Ames  & Ames mutagenicity & 6512 & \cite{ames_source} \\
 Bace  & $\beta$-Secretase 1 inhibition & 1513 & \cite{bace_source} \\
 BBBP  & Blood–brain barrier penetration & 2039 & \cite{bbbp_source} \\
 Beet  & Toxicity in honeybees & 254 & \cite{beet_source} \\
 ClinTox  & Clinical trial results & 1478 & \cite{clintox_source} \\
 \hline
 
 \end{tabular}
\caption{\label{tab:datasets} The five classification data sets used in benchmarking our models.}
\end{table}

\subsection{Hyperparameter Selection} \label{Parameters}

In this section, we outline our selections for the various parameters needed for the models in the present work. The transformer model from \cite{hundreds} requires specification of a pretraining data set. Chen et al. generated their proposed self-supervised learning platform fingerprints (SSLP-FPs) using pretraining data from the publicly available ChEMBL \cite{chembl-source}, PubChem \cite{pubchem-source}, and ZINC \cite{zinc-source} data sets. In the reporting of their results, as in Table \ref{tab:90 Comparison}, the pretraining data set used for each set of SSLP-FPs is recorded. Set C indicates only the ChEMBL data set, CP indicates the union of the ChEMBL and PubChem data sets, and CPZ indicates the union of all three aforementioned data sets. Notably, in \cite{hundreds}, of the C, CP, and CPZ models, the C transformer model (i.e., the model pretrained on only the ChEBML data set) achieved the best predictive performance for three of the five benchmark data sets used in the present work. Consequently, we use only the ChEMBL data set as our pretraining set. Aside from the model selection for the transformer, the generation of AE-FPs and BT-FPs did not require any parameter selection on our part; in fact, all model parameters were saved in the pretrained models \cite{hundreds, gao2021proteome}.

To generate ECFPs, we used the same parameters as outlined by Chen et al. \cite{hundreds} with the Morgan fingerprint algorithm from the RDKit library \cite{rdkit}. That is, we used three radii (1, 2, and 3) and three folding lengths (512, 1024, and 2048) for a total of nine parameter settings, which are indicated in the tables and plots in Section \ref{Performance}. Note that the ECFPs generated using a radius of 1, 2, or 3 are denoted by 2, 4, and 6, respectively, in the table to account for the discrepancy between the RDKit Morgan algorithm and ECFP algorithm conventions (see Section \ref{ecfps} for more details). 

Our MBO algorithm (outlined in Algorithm \ref{alg-mbo}) requires several parameters, some of which we set manually and some of which we designated as hyperparameters to tune. The MBO algorithm begins by constructing a graph with $N_n$ neighbors for each data point (as opposed to constructing the full graph). After constructing the graph and computing the graph Laplacian, we compute its first $N_e$ eigenavlues and eigenvectors. We designated both $N_n$ and $N_e$ as hyperparameters to tune. Additional hyperparameters for the MBO algorithm are the factor $C$ in the diffusion operator, the time step $dt$ in applying the diffusion operator, and the maximum number of iterations $N_t$ of the diffusion and thresholding steps. We searched the space of these parameters and tuned to yield the best ROC-AUC score for each data set with a given set of features and amount of labeled data. The optimal hyperparameters for our proposed AE-MBO, BT-MBO, and ECFP-MBO methods are detailed in the Supporting Information.

The number of times $N_s$ to apply the diffusion operator in the MBO algorithm was set to 3 for all tasks. The number of labeled points or points in the training set $N_p$ was varied for different tasks; we conducted experiments with 1\% up to 90\% of the points in each data set being labeled. In the cases of 1\%, 2\%, and 5\% labeled data, we set the number of labeled sets $N_l$ to 50. For all other amounts of labeled data, we set the number of labeled sets $N_l$ to 10. 

As detailed in Section \ref{Performance}, we compare the performance of our proposed methods to the methods proposed and cited in \cite{hundreds}, which use gradient boosting decision tree (GBDT), random forest (RF), and support vector machine (SVM) algorithms for their downstream classification tasks. We also generate some of our own comparisons using GBDT, RF, and SVM algorithms. To appropriately compare results across these methods, we set all parameters used in these algorithms to be identical to those used by Chen et al., which are listed in the Supporting Information in \cite{hundreds}.

\subsection{Evaluation and Visualization Metrics} \label{evaluation}

In all classification tasks, model performance is quantified using the area under the receiver operating characteristic (ROC) curve, or the ROC-AUC score. {\color{black} Given predicted labels from a classifier for a set of data along with the associated true labels,} the ROC curve plots the true positive rate (TPR) versus the false positive rate (FPR) of the classifier. Given a classification task with positive and negative labels, the TPR is defined as the ratio of true positives (i.e., correctly classified positive data points) to the total number of actual positive data points. The FPR is defined as the ratio of false positives (i.e., negative data points that the classifier predicted as positive) to the total number of negative data points. A higher ROC-AUC score indicates that the classifier is better at distinguishing between the two classes \cite{hundreds}. {\color{black} For all proposed MBO-integrated models in the present work, we calculated the ROC-AUC score using the \texttt{perfcurve} function in MATLAB. For the comparison methods using RF, GBDT, and SVM algorithms, we calculated the ROC-AUC score using the \texttt{roc\_auc\_score} function from the scikit-learn library in Python.}

To further examine the various featurizations used for the data sets in the present work, we also utilized the residue-similarity (R-S) scores and corresponding R-S plot proposed by Hozumi et al. in \cite{ccp}. Consider a classification problem with labeled data $\mathscr{L} = \{(\mathbf{x}_i, y_i)\}_{i=1}^N$, where $y_i \in \{0,1,\dots, K - 1\}$ is the ground truth label of the data point $\mathbf{x}_i$, $N$ is the number of samples, and $K$ is the number of classes. Let $C_k = \{\mathbf{x}_i |y_i = k\} $. Then, the residue score for $\mathbf{x}_i$ is defined as
\begin{equation*}
    R_i := R(\mathbf{x}_i) = \frac{1}{R_\text{max} + \epsilon}\sum_{\mathbf{x}_j \notin C_k}\| \mathbf{x}_i - \mathbf{x}_j\|,
\end{equation*}
where $\| \cdot \|$ is Euclidean distance between vectors, and $R_\text{max}$ is the maximal residue score over all data points. The similarity score for $\mathbf{x}_i$ is defined as
\begin{equation*}
    S_i := S(\mathbf{x}_i) = \frac{1}{|C_k| + \epsilon}\sum_{\mathbf{x}_j \in C_k} \left(1 - \frac{\| \mathbf{x}_i - \mathbf{x}_j\|}{d_\text{max} + \epsilon} \right),
\end{equation*}
where $d_\text{max}$ is the maximal pairwise distance of the data set. In both defined quantities above, $\epsilon$ is a small constant intended to account for cases in which $R_\text{max}$, $|C_k|$, and $d_\text{max}$ may be equal to zero, which can occur when some $C_k$ is empty for a given training/testing split of a data set. Because our focus is on very small training sets, certain random training/testing splits of our data sets can result in training sets containing data points from only one class. To make such situations computationally stable, we set $\epsilon = 10^{-5}$ in the present work.

Both the residue and similarity scores for a data point have a range of $[0,1]$ due to scaling. The residue score measures the sum of distances across classes; that is, a high residue score indicates that a class is far from other classes. The similarity score measures the average score within a class; consequently, a high similarity score indicates that data in a class is clustered close together \cite{ccp}. The R-S plot graphs the similarity score versus the residue score for each data point. In the plots used in the present work, R-S plots are separated by class, and each data point in the R-S plot is assigned a color indicating its predicted label. Further details are given in Section \ref{Performance} along with selected R-S plots.

The residue and similarity scores discussed above are defined for single data points. The class residue index (CRI) and class similarity index (CSI) can be defined for an entire class $k$ as $$\text{CRI}_k = \frac{1}{|C_k|+ \epsilon}\sum_i R_i$$ and $$\text{CSI}_k = \frac{1}{|C_k|+ \epsilon}\sum_i S_i,$$ respectively. These indices allow for comparisons across classes \cite{ccp}.

Furthermore, indices independent of classes can be defined, allowing for description of an entire data set. Two such indices are the residue index (RI), defined as $$\text{RI} = \frac{1}{K}\sum_k \text{CRI}_k,$$ and the similarity index (SI), defined as $$\text{SI} = \frac{1}{K}\sum_k \text{CSI}_k.$$ Higher residue and similarity indices are better for a data set. A high residue index indicates that the classes or clusters in a data set are well separated on average, and a high similarity index indicates that the data points in each class tend to be similar or clustered close together. Finally, we can define R-S index (RSI) as \cite{ccp} $$\text{RSI} = 1 - |\text{RI} - \text{SI}|.$$  Note that, by definition, all of these indices have a range of $[0,1]$. Visualizations of the R-S indices and selected R-S scores associated with all three models on all five data sets are displayed in Section \ref{Performance}.

\subsection{Performance and Discussion} \label{Performance}

To examine the effectiveness of MBO-based models for low amounts of labeled data compared to those involving other machine learning algorithms, we evaluated our models' performance against other methods for 1\%, 2\%, 5\%, and 90\% labeled data passed to the MBO model or other machine learning algorithm for the five molecular classification data sets. To generate comparisons for 1\%, 2\%, and 5\% labeled data, we reproduced the processes proposed by Chen et al. \cite{hundreds} for the lower percentages of labeled data. In this section, we compare the performance of our proposed BT-MBO, AE-MBO, ECFP-MBO, and Consensus-MBO models to that of the transformer and autoencoder model \cite{hundreds, gao2021proteome} paired with gradient boosting decision tree (BT-GBDT and AE-GBDT), random forest (BT-RF and AE-RF), and support vector machine (BT-SVM and AE-SVM) algorithms, as well as the ECFP algorithm \cite{circ-fp, rdkit} paired with GBDT, RF, and SVM. 

More specifically, for the BT-GBDT, BT-RF, and BT-SVM models, we pass the BT-FPs generated from the transformer model to GBDT, RF, and SVM algorithms, respectively. For the AE-GBDT, AE-RF, and AE-SVM models, we pass the AE-FPs generated from the autoencoder model to the GBDT, RF, and SVM algorithms, respectively, for prediction. The results from these six models for 1\%, 2\%, and 5\% labeled data are shown in Tables \ref{tab:1 Comparison}, \ref{tab:2 Comparison}, and \ref{tab:5 Comparison}. To evaluate the ECFP algorithm performance with other machine learning algorithms, we pass the ECFPs generated by each of the nine parameter settings as features to the GBDT, RF, and SVM algorithms. We chose the results from the best-performing machine learning algorithm for each parameter setting and data set to display in Tables \ref{tab:1 Comparison}, \ref{tab:2 Comparison}, and \ref{tab:5 Comparison}. In all cases, we use the GBDT, RF, and SVM algorithms from the scikit-learn library. Note that in all classification tasks, the models are evaluated using the ROC-AUC score (see Section \ref{evaluation} for more details).

The comparisons for 90\% labeled data are reported slightly differently, as all results for this percentage of labeled data in Table \ref{tab:90 Comparison} other than those for our proposed methods were generated and reported by Chen et al. \cite{hundreds}. In particular, note that the transformer models listed in the table are SSLP-FPs(C), SSLP-FPs(CP), and SSLP-FPs(CPZ), which are the self-supervised learning platform fingerprints generated from their transformer model with three different pretraining sets. For the prediction task for each of the three models, Chen et al. used GBDT, RF, and SVM algorithms from the scikit-learn library, then chose the best-performing of the three algorithms for their final results. Similarly, for the autoencoder model performance (Auto-FPs), they chose the best-performing of the GBDT, RF, and SVM algorithms. The results for the ECFP algorithm are reported in the same manner as in Tables \ref{tab:1 Comparison}, \ref{tab:2 Comparison}, and \ref{tab:5 Comparison}. For each listed model, Chen et al. evaluated its performance using 10-fold cross-validation. Specifically, they randomly split the five data sets 10 times into training, validation, and testing sets in the ratio of 8:1:1, for 90\% labeled data and 10\% unlabeled data. The results are given as the average ROC-AUC score for the 10 splits.

Similarly, to evaluate our proposed BT-MBO, AE-MBO, and ECFP-MBO methods for 90\% labeled data, we {\color{black} randomly} split each input data set into labeled and unlabeled sets 10 times and performed the MBO method for each split. {\color{black} To split the data into labeled and unlabeled sets with $N_p$ labeled points for the MBO scheme, we randomly chose $N_p$ points of the input data set to be labeled and designated the remaining data points as unlabeled. The known labels were then propagated on the graph to the unlabeled elements, as described in Section \ref{MBO} as part of the MBO scheme.} For each of the 10 splits, {\color{black} after performing the MBO method to generate predicted labels,} we calculated the ROC-AUC score {\color{black} for that split. We} then averaged the ROC-AUC scores {\color{black} of the 10 splits}. These results are displayed in Table \ref{tab:90 Comparison}. For 1\%, 2\%, and 5\% labeled data, we repeated this process with 50 instead of 10 training/testing splits and calculated the average ROC-AUC score over the 50 data splits. In these cases with low percentages of labeled data, the training sets contained very few points, so the increase in the number of different data partitions was intended to account for the likely variability among partitions. All displayed results for the proposed methods are the best average ROC-AUC score over all tested parameters (see Section \ref{Parameters} for information about parameter search and selection).

To evaluate our Consensus-MBO method, we chose the two top-performing methods among BT-MBO, AE-MBO, and ECFP-MBO for a given data set and percentage of labeled data. We then applied the Consensus-MBO model (more details in \ref{Consensus}) as described above, with 10 labeled sets for 90\% labeled data and 50 labeled sets for 1\%, 2\%, and 5\% labeled data. We repeated the Consensus-MBO method in this way for 10 total trials. The results recorded in Tables \ref{tab:1 Comparison}, \ref{tab:2 Comparison}, \ref{tab:5 Comparison}, and \ref{tab:90 Comparison} for the Consensus-MBO method are the average ROC-AUC score and standard deviation over these 10 trials. 

\begin{figure}
    \centering
    \begin{minipage}{0.5\textwidth}
        \centering
        \includegraphics[width=\textwidth]{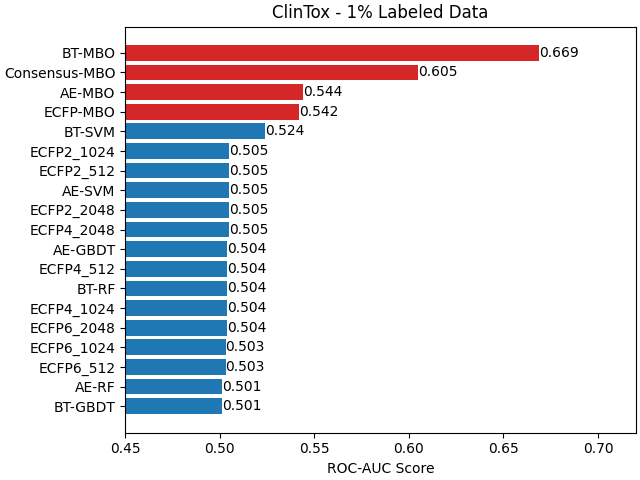}
    \end{minipage}\hfill
    \begin{minipage}{0.5\textwidth}
        \centering
        \includegraphics[width=\textwidth]{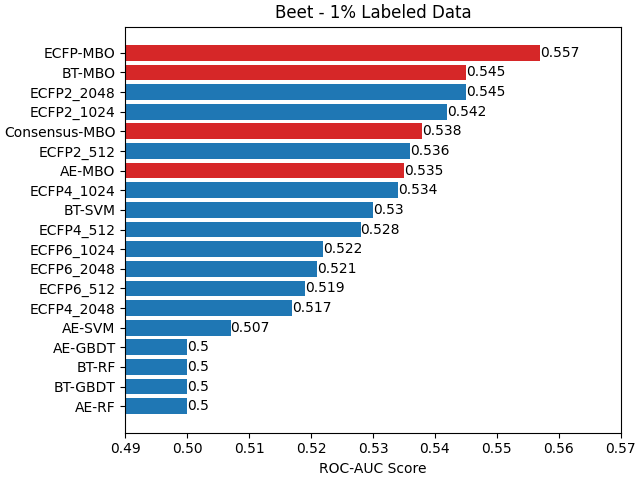}
    \end{minipage}
    \begin{minipage}{0.5\textwidth}
        \centering
        \includegraphics[width=\textwidth]{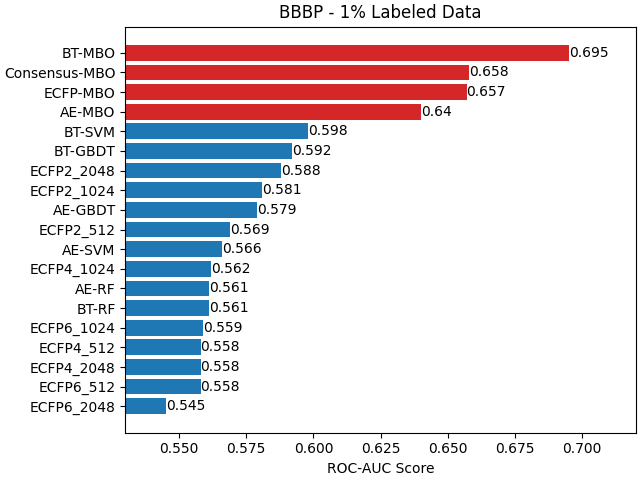}
    \end{minipage}\hfill
    \begin{minipage}{0.5\textwidth}
        \centering
        \includegraphics[width=\textwidth]{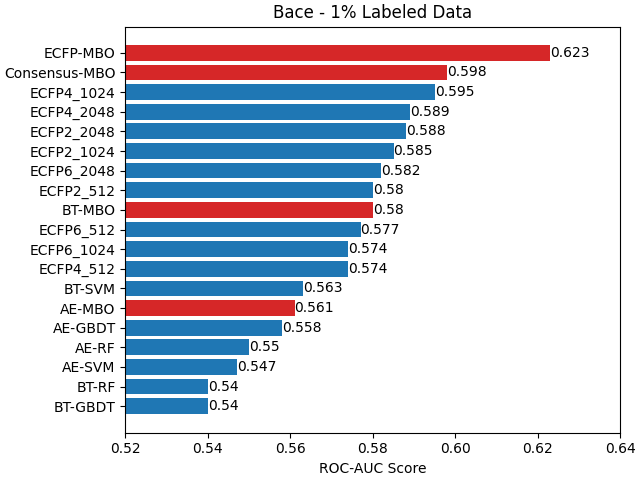}
    \end{minipage}
    \includegraphics[width=0.5\textwidth]{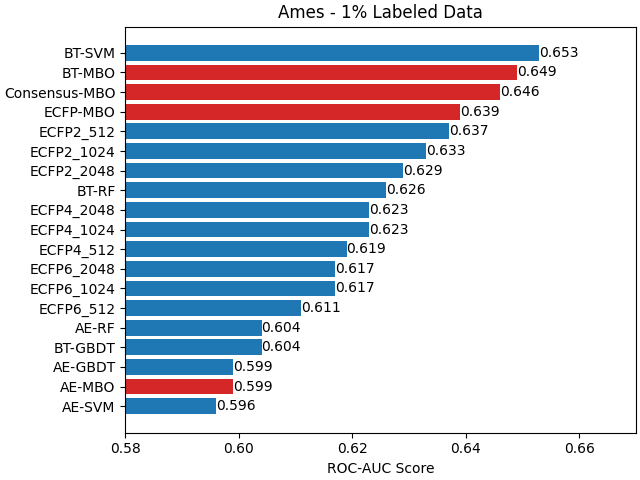}
    \caption{Comparison of our proposed methods (shown in red) with other methods (shown in blue) on the five benchmark data sets for 1\% labeled data.}
    \label{1pfig}
\end{figure}

Table \ref{tab:1 Comparison} and Figure \ref{1pfig} show the results for 1\% labeled data for the five molecular classification data sets. For four out of five data sets, one of our proposed models outperforms the best-performing comparison methods. In all cases, either the BT-MBO or ECFP-MBO method was the best performing of our methods. Note that the Consensus-MBO method did not yield the best results for any data set, likely due to the relatively large difference in performance between our two best-performing methods for most of the data sets. With 1\% labeled data, our models perform particularly well compared to other methods for the ClinTox data set, with our BT-MBO method achieving an ROC-AUC score of 0.669 versus the best score of 0.524 for other methods, and for the BBBP data set, with our BT-MBO model achieving an ROC-AUC score of 0.695 versus 0.598 for other methods. 

\begin{figure}
    \centering
    \begin{minipage}{0.5\textwidth}
        \centering
        \includegraphics[width=\textwidth]{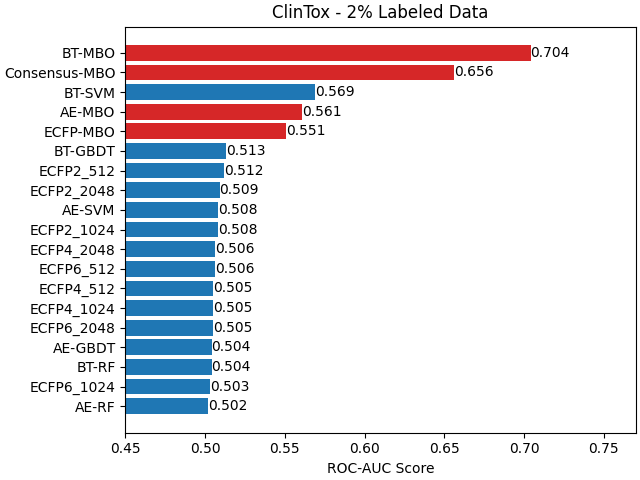}
    \end{minipage}\hfill
    \begin{minipage}{0.5\textwidth}
        \centering
        \includegraphics[width=\textwidth]{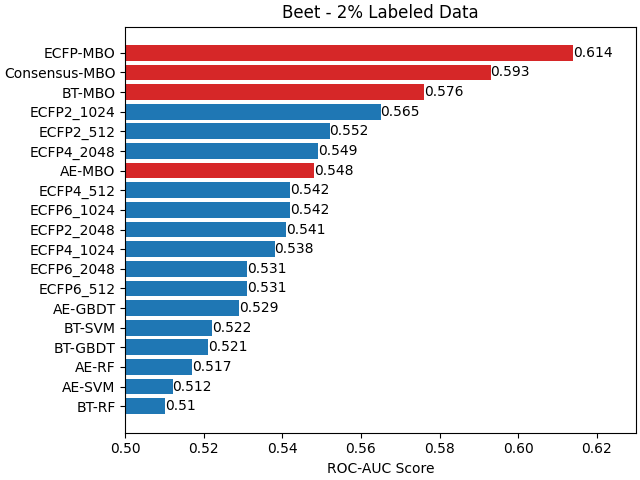}
    \end{minipage}
    \begin{minipage}{0.5\textwidth}
        \centering
        \includegraphics[width=\textwidth]{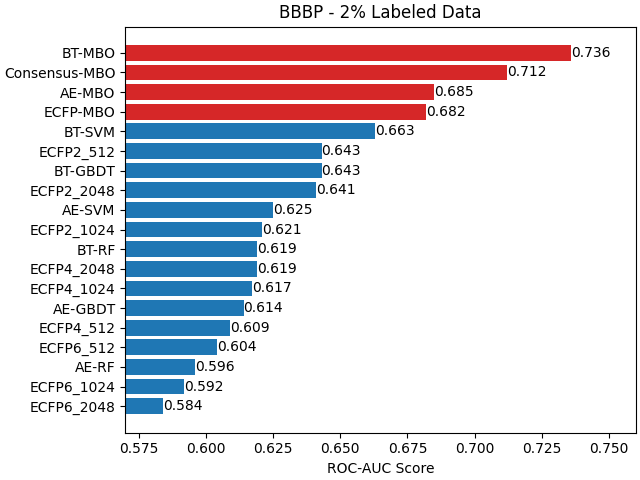}
    \end{minipage}\hfill
    \begin{minipage}{0.5\textwidth}
        \centering
        \includegraphics[width=\textwidth]{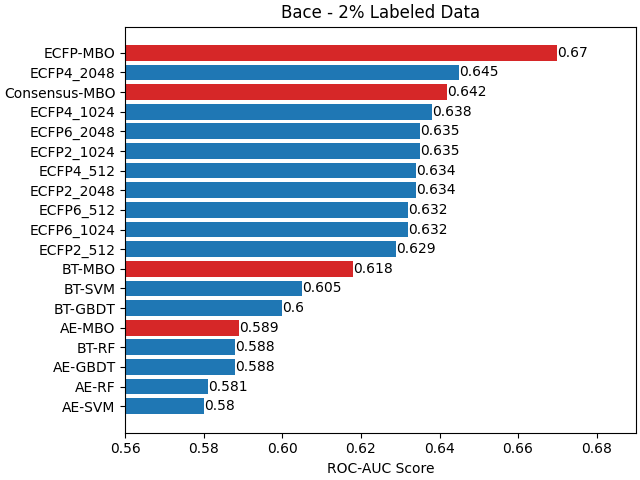}
    \end{minipage}
    \includegraphics[width=0.5\textwidth]{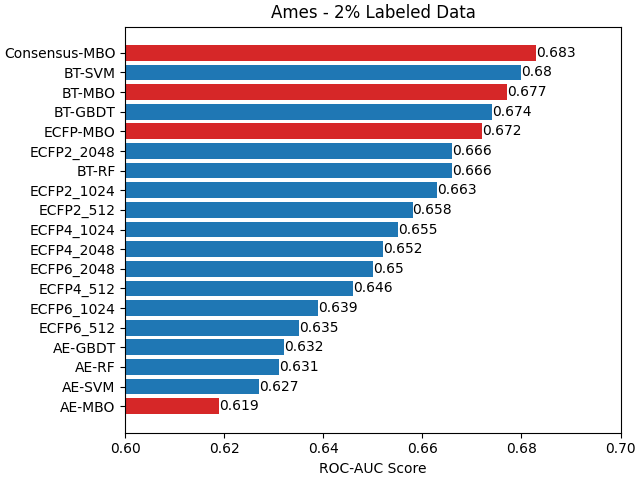}
    \caption{Comparison of our proposed methods (shown in red) with other methods (shown in blue) on the five benchmark data sets for 2\% labeled data.}
    \label{2pfig}
\end{figure}

The results for 2\% labeled data are displayed in Table \ref{tab:2 Comparison} and Figure \ref{2pfig}. Here, one of our proposed models outperformed all other comparison methods for all five data sets. Again, our models perform significantly better than other methods on the ClinTox and BBBP sets, with our BT-MBO method achieving an ROC-AUC score of 0.704 for the ClinTox data set and a score of 0.736 for the BBBP data set. The best performance achieved by the comparison methods was an ROC-AUC score of 0.569 for the ClinTox data set and 0.663 for the BBBP data set, both from the BT-SVM method. Our proposed ECFP-MBO method achieved the best performance for the Beet (ROC-AUC = 0.614) and Bace (ROC-AUC = 0.67) data sets. For the Ames data set, our Consensus-MBO method achieved the best performance (ROC-AUC = 0.683), averaging our BT-MBO and ECFP-MBO methods.

\begin{table}[h!]
\centering
\makebox[\textwidth]{
\begin{tabular}{c|c c c c c}
\hline
&\multicolumn{5}{c}{\textbf{ROC-AUC Scores for 1\% Labeled Data}}\\
\textbf{Model} & \textbf{Ames} & \textbf{Bace} & \textbf{BBBP} & \textbf{Beet} & \textbf{ClinTox}\\
\hline
BT-MBO (Proposed) & $0.649 \pm .024$ & $0.580 \pm .039$ & $\mathbf{0.695 \pm .068}$ & $0.545 \pm .062$ & $\mathbf{0.669 \pm .123}$ \\
AE-MBO (Proposed) & $0.600 \pm .023$ & $0.561 \pm .038$ & $0.640 \pm .053$ & $0.535 \pm .048$ & $0.544 \pm .038$\\
ECFP-MBO (Proposed) & $0.639 \pm .030$ & $\mathbf{0.623 \pm .058}$ & $0.657 \pm .044$ & $\mathbf{0.557 \pm .064}$ & $0.542 \pm .032$\\
Consensus-MBO (Proposed) & $0.646 \pm .033$ & $0.598 \pm .049$ & $0.658 \pm .055$ & $0.538 \pm .066$ & $0.605 \pm .094$\\
BT-GBDT \cite{hundreds} & $0.604 \pm .036$ & $0.540 \pm .029$ & $0.592 \pm .074$ & $0.5 \pm 0.0$ & $0.501 \pm .003$\\
BT-RF \cite {hundreds} & $0.626 \pm .031$ & $0.540 \pm .034$ & $0.561 \pm .070$ & $0.5 \pm 0.0$ & $0.504 \pm .019$\\
BT-SVM \cite {hundreds} & $\mathbf{0.653 \pm .027}$ & $0.563 \pm .035$ & $0.598 \pm .096$ & $0.530 \pm .053$ & $0.524 \pm .037$\\
AE-GBDT \cite{gao2021proteome} & $0.599 \pm .028$ & $0.558 \pm .035$ & $0.579 \pm .061$ & $0.5 \pm 0.0$ & $0.504 \pm .008$ \\
AE-RF \cite{gao2021proteome} & $0.604 \pm .027$ & $0.550 \pm .038$ & $0.561 \pm .060$ & $0.5 \pm 0.0$ & $0.501 \pm .003$ \\
AE-SVM \cite{gao2021proteome} & $0.596 \pm .024$ & $0.547 \pm .036$ & $0.566 \pm .066$ & $0.507 \pm .045$ & $0.505 \pm .008$ \\
ECFP2\_512 \cite {circ-fp} & $0.637 \pm .028$ & $0.580 \pm .044 $ & $0.569 \pm .050$ & $0.536 \pm .049$ & $0.505 \pm .007$\\
ECFP2\_1024 \cite {circ-fp} & $0.633 \pm .027$ & $0.585 \pm .040$ & $0.581 \pm .060$ & $0.542 \pm .058$ & $0.505 \pm .008$\\
ECFP2\_2048 \cite {circ-fp} & $0.629 \pm .028$ & $0.588 \pm .040$ & $0.588 \pm .063$ & $0.545 \pm .072$ & $0.505 \pm .008$\\
ECFP4\_512 \cite {circ-fp} & $0.619 \pm .029$ & $0.574 \pm .052$ & $0.558 \pm .048$ & $0.528 \pm .049$ & $0.504 \pm .008$\\
ECFP4\_1024 \cite {circ-fp} & $0.623 \pm .030$ & $0.595 \pm .041$ & $0.562 \pm .051$ & $0.534 \pm .056$ & $0.504 \pm .005$\\
ECFP4\_2048 \cite {circ-fp} & $0.623 \pm .028$ & $0.589 \pm .048$ & $0.558 \pm .048$ & $0.517 \pm .034$ & $0.505 \pm .006$\\
ECFP6\_512 \cite {circ-fp} & $0.611 \pm .026$ & $0.577 \pm .041$ & $0.558 \pm .054$ & $0.519 \pm .038$ & $0.503 \pm .006$\\
ECFP6\_1024 \cite {circ-fp} & $0.617 \pm .024$ & $0.574 \pm .045$ & $0.559 \pm .045$ & $0.522 \pm .036$ & $0.503 \pm .005$\\
ECFP6\_2048 \cite {circ-fp} & $0.617 \pm .028$ & $0.582 \pm .045$ & $0.545 \pm .042$ & $0.521 \pm .042$ & $0.504 \pm .005$\\

\end{tabular}}
\caption{\label{tab:1 Comparison} Comparison of our models' performance on the five data sets with 1\% labeled data with other models' performance. Our four models are AE-MBO (autoencoder with MBO), BT-MBO (transformer with MBO), ECFP-MBO (extended-connectivity fingerprints with MBO), and Consensus-MBO, a model generating the consensus from our top two scoring methods for a given data set and percent of labeled data (more details in Section \ref{Consensus}). The BT-GBDT, BT-RF, and BT-SVM models use the BT-FPs as features for gradient boosting decision trees, random forest, and support vector machine, respectively (note that these models are denoted by 'SSLP-FP' in \cite{hundreds}). AE-GBDT, AE-RF, and AE-SVM refer to the AE-FPs used as features for the specified machine learning method. Performance is given as average ROC-AUC score over 50 labeled sets with standard deviation, and Consensus-MBO performance is given as the average over 10 trials with standard deviation.}
\end{table}

\begin{table}[h!]
\centering
\makebox[\textwidth]{
 \begin{tabular}{c|c c c c c}
 \hline
 &\multicolumn{5}{c}{\textbf{ROC-AUC Scores for 2\% Labeled Data}}\\
 \textbf{Model} & \textbf{Ames} & \textbf{Bace} & \textbf{BBBP} & \textbf{Beet} & \textbf{ClinTox}\\
 \hline
 BT-MBO (Proposed) & $0.677 \pm .021$ & $0.618 \pm .037$ & $\mathbf{0.736 \pm .051}$ & $0.576 \pm .075$ & $\mathbf{0.704 \pm .113}$ \\
 AE-MBO (Proposed) & $0.619 \pm .016$ & $0.589 \pm .029$ & $0.685 \pm .048$ & $0.548 \pm .067$ & $0.561 \pm .030$ \\
 ECFP-MBO (Proposed) & $0.672 \pm .021$ & $\mathbf{0.670 \pm .034}$ & $0.682 \pm .028$ & $\mathbf{0.614 \pm .089}$ & $0.551 \pm .026$\\
 Consensus-MBO (Proposed) & $\mathbf{0.683 \pm .023}$ & $0.642 \pm .043$ & $0.712 \pm .058$ & $0.593 \pm .090$ & $0.656 \pm .076$\\
 BT-GBDT \cite{hundreds} & $0.674 \pm .023$ & $0.600 \pm .036$ & $0.643 \pm .075$ & $0.521 \pm .035$ & $0.513 \pm .024$ \\
 BT-RF \cite {hundreds} & $0.666 \pm .025$ & $0.588 \pm .034$ & $0.619 \pm .071$ & $0.510 \pm .020$ & $0.504 \pm .010$ \\
 BT-SVM \cite {hundreds} & $0.680 \pm .017$ & $0.605 \pm .036$ & $0.663 \pm .082$ & $0.522 \pm .040$ & $0.569 \pm .080$ \\
 AE-GBDT \cite{gao2021proteome} & $0.632 \pm .018$ & $0.588 \pm .038$ & $0.614 \pm .068$ & $0.529 \pm .036$ & $0.504 \pm .010$ \\
 AE-RF \cite{gao2021proteome} & $0.631 \pm .019$ & $0.581 \pm .034$ & $0.596 \pm .054$ & $0.517 \pm .022$ & $0.502 \pm .004$ \\
 AE-SVM \cite{gao2021proteome} & $0.627 \pm .015$ & $0.580 \pm .035$ & $0.625 \pm .066$ & $0.512 \pm .028$ & $0.508 \pm .012$ \\
 ECFP2\_512 \cite {circ-fp} & $0.658 \pm .018$ & $0.629 \pm .039$ & $0.643 \pm .051$ & $0.552 \pm .064$ & $0.512 \pm .017$\\
 ECFP2\_1024 \cite {circ-fp} & $0.663 \pm .018$ & $0.635 \pm .038$ & $0.621 \pm .048$ & $0.565 \pm .057$ & $0.508 \pm .010$\\
 ECFP2\_2048 \cite {circ-fp} & $0.666 \pm .019$ & $0.634 \pm .044$ & $0.641 \pm .041$ & $0.541 \pm .058$ & $0.509 \pm .010$\\
 ECFP4\_512 \cite {circ-fp} & $0.646 \pm .019$ & $0.634 \pm .042$ & $0.609 \pm .052$ & $0.542 \pm .052$ & $0.505 \pm .008$\\
 ECFP4\_1024 \cite {circ-fp} & $0.655 \pm .020$ & $0.638 \pm .036$ & $0.617 \pm .045$ & $0.538 \pm .050$ & $0.505 \pm .006$\\
 ECFP4\_2048 \cite {circ-fp} & $0.652 \pm .021$ & $0.645 \pm .040$ & $0.619 \pm .053$ & $0.549 \pm .057$ & $0.506 \pm .008$\\
 ECFP6\_512 \cite {circ-fp} & $0.635 \pm .025$ & $0.632 \pm .040$ & $0.604 \pm .051$ & $0.531 \pm .056$ & $0.506 \pm .011$\\
 ECFP6\_1024 \cite {circ-fp} & $0.639 \pm .020$ & $0.632 \pm .046$ & $0.592 \pm .048$ & $0.542 \pm .049$ & $0.503 \pm .006$\\
 ECFP6\_2048 \cite {circ-fp} & $0.650 \pm .022$ & $0.635 \pm .046$ & $0.584 \pm .046$ & $0.531 \pm .047$ & $0.505 \pm .007$\\

 \end{tabular}}
\caption{\label{tab:2 Comparison} Comparison of our models' performance on the five data sets with 2\% labeled data with other models' performance. Our four models are AE-MBO (autoencoder with MBO), BT-MBO (transformer with MBO), ECFP-MBO (extended-connectivity fingerprints with MBO), and Consensus-MBO, a model generating the consensus from our top two scoring methods for a given data set and percent of labeled data (more details in Section \ref{Consensus}). The BT-GBDT, BT-RF, and BT-SVM models use the BT-FPs as features for gradient boosting decision trees, random forest, and support vector machine, respectively (note that these models are denoted by 'SSLP-FP' in \cite{hundreds}). AE-GBDT, AE-RF, and AE-SVM refer to the AE-FPs used as features for the specified machine learning method. Performance is given as average ROC-AUC score over 50 labeled sets with standard deviation, and Consensus-MBO performance is given as the average over 10 trials with standard deviation.}
\end{table}

\begin{table}[!h]
\centering
\makebox[\textwidth]{
\begin{tabular}{c|c c c c c}
\hline
&\multicolumn{5}{c}{\textbf{ROC-AUC Scores for 5\% Labeled Data}}\\
\textbf{Model} & \textbf{Ames} & \textbf{Bace} & \textbf{BBBP} & \textbf{Beet} & \textbf{ClinTox}\\
\hline
BT-MBO (Proposed) & $0.716 \pm .014$ & $0.680 \pm .027$ & $\mathbf{0.785 \pm .036}$ & $0.621 \pm .070$ & $\mathbf{0.774 \pm .058}$ \\
AE-MBO (Proposed) & $0.653 \pm .012$ & $0.646 \pm .024$ & $0.730 \pm .040$ & $0.578 \pm .047$ & $0.596 \pm .032$ \\
ECFP-MBO (Proposed) & $0.710 \pm .012$ & $\mathbf{0.720 \pm .026}$ & $0.721 \pm .025$ & $0.662 \pm .073$ & $0.563 \pm .029$\\
Consensus-MBO (Proposed) & $\mathbf{0.722 \pm .013}$ & $0.702 \pm .032$ & $0.765 \pm .034$ & $\mathbf{0.666 \pm .070}$ & $0.712 \pm .055$\\
BT-GBDT \cite{hundreds} & $0.717 \pm .009$ & $0.654 \pm .029$ & $0.696 \pm .061$ & $0.551 \pm .054$ & $0.524 \pm .029$ \\
BT-RF \cite {hundreds} & $0.709 \pm .014$ & $0.642 \pm .030$ & $0.684 \pm .057$ & $0.555 \pm .058$ & $0.513 \pm .025$ \\
BT-SVM \cite {hundreds} & $0.721 \pm .011$ & $0.679 \pm .027$ & $0.739 \pm .052$ & $0.566 \pm .051$ & $0.635 \pm .084$ \\
AE-GBDT \cite{gao2021proteome} & $0.666 \pm .011$ & $0.653 \pm .030$ & $0.665 \pm .050$ & $0.549 \pm .042$ & $0.506 \pm .007$ \\
AE-RF \cite{gao2021proteome} & $0.662 \pm .013$ & $0.663 \pm .028$ & $0.632 \pm .063$ & $0.551 \pm .040$ & $0.503 \pm .006$ \\
AE-SVM \cite{gao2021proteome} & $0.653 \pm .010$ & $0.644 \pm .028$ & $0.716 \pm .044$ & $0.535 \pm .036$ & $0.520 \pm .023$ \\
ECFP2\_512 \cite {circ-fp} & $0.700 \pm .010$ & $0.706 \pm .025$ & $0.703 \pm .029$ & $0.596 \pm .059$ & $0.517 \pm .018$\\
ECFP2\_1024 \cite {circ-fp} & $0.703 \pm .011$ & $0.701 \pm .031$ & $0.701 \pm .028$ & $0.603 \pm .057$ & $0.513 \pm .010$\\
ECFP2\_2048 \cite {circ-fp} & $0.705 \pm .009$ & $0.700 \pm .031$ & $0.696 \pm .033$ & $0.609 \pm .063$ & $0.512 \pm .012$\\
ECFP4\_512 \cite {circ-fp} & $0.685 \pm .015$ & $0.706 \pm .025$ & $0.677 \pm .036$ & $0.575 \pm .058$ & $0.513 \pm .011$\\
ECFP4\_1024 \cite {circ-fp} & $0.691 \pm .010$ & $0.704 \pm .032$ & $0.686 \pm .036$ & $0.587 \pm .050$ & $0.515 \pm .012$\\
ECFP4\_2048 \cite {circ-fp} & $0.700 \pm .013$ & $0.712 \pm .025$ & $0.678 \pm .032$ & $0.602 \pm .064$ & $0.510 \pm .010$\\
ECFP6\_512 \cite {circ-fp} & $0.669 \pm .012$ & $0.694 \pm .029$ & $0.664 \pm .032$ & $0.580 \pm .057$ & $0.508 \pm .009$\\
ECFP6\_1024 \cite {circ-fp} & $0.677 \pm .009$ & $0.698 \pm .025$ & $0.672 \pm .033$ & $0.574 \pm .055$ & $0.511 \pm .009$\\
ECFP6\_2048 \cite {circ-fp} & $0.688 \pm .012$ & $0.710 \pm .028$ & $0.670 \pm .034$ & $0.572 \pm .057$ & $0.510 \pm .009$\\

\end{tabular}}
\caption{\label{tab:5 Comparison} Comparison of our models' performance on the five data sets with 5\% labeled data with other models' performance. Our four models are AE-MBO (autoencoder with MBO), BT-MBO (transformer with MBO), ECFP-MBO (extended-connectivity fingerprints with MBO), and Consensus-MBO, a model generating the consensus from our top two scoring methods for a given data set and percent of labeled data (more details in Section \ref{Consensus}). The BT-GBDT, BT-RF, and BT-SVM models use the BT-FPs as features for gradient boosting decision trees, random forest, and support vector machine, respectively (note that these models are denoted by 'SSLP-FP' in \cite{hundreds}). AE-GBDT, AE-RF, and AE-SVM refer to the AE-FPs used as features for the specified machine learning method. Performance is given as average ROC-AUC score over 50 labeled sets with standard deviation, and Consensus-MBO performance is given as the average over 10 trials with standard deviation.}
\end{table}

\begin{figure}
    \centering
    \begin{minipage}{0.5\textwidth}
        \centering
        \includegraphics[width=\textwidth]{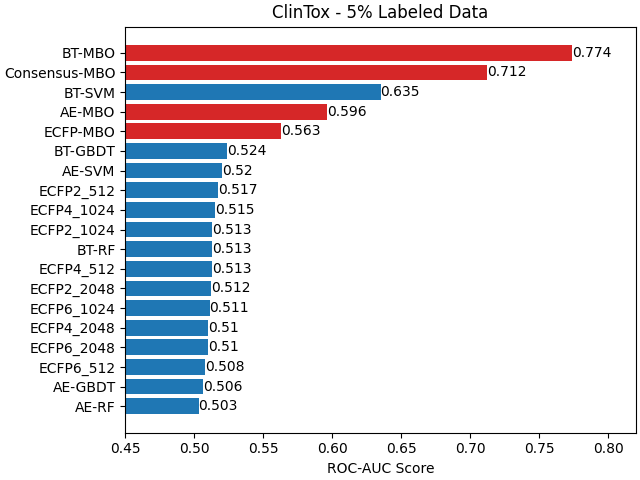}
    \end{minipage}\hfill
    \begin{minipage}{0.5\textwidth}
        \centering
        \includegraphics[width=\textwidth]{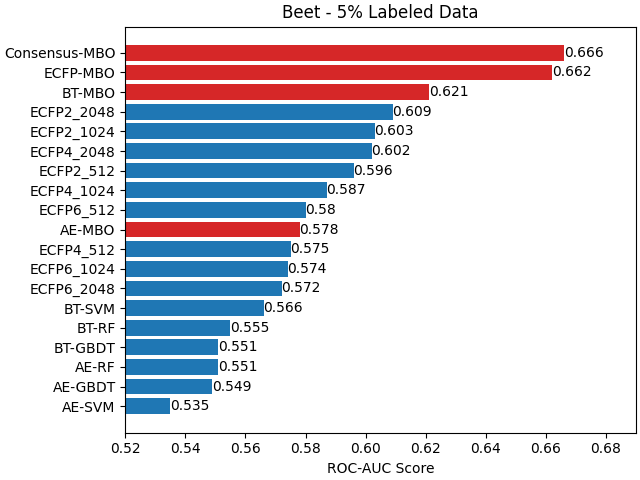}
    \end{minipage}
    \begin{minipage}{0.5\textwidth}
        \centering
        \includegraphics[width=\textwidth]{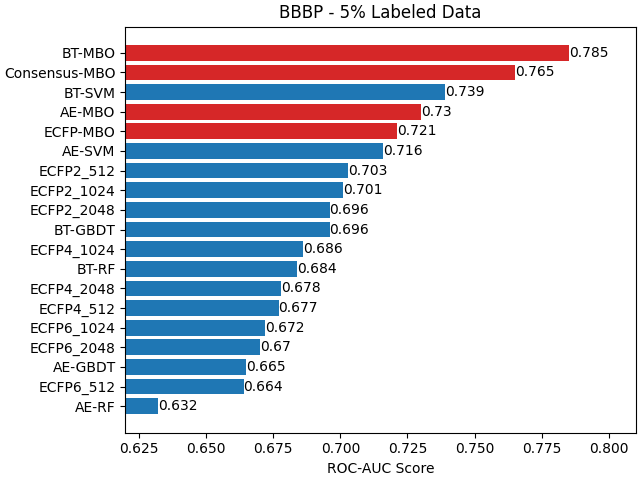}
    \end{minipage}\hfill
    \begin{minipage}{0.5\textwidth}
        \centering
        \includegraphics[width=\textwidth]{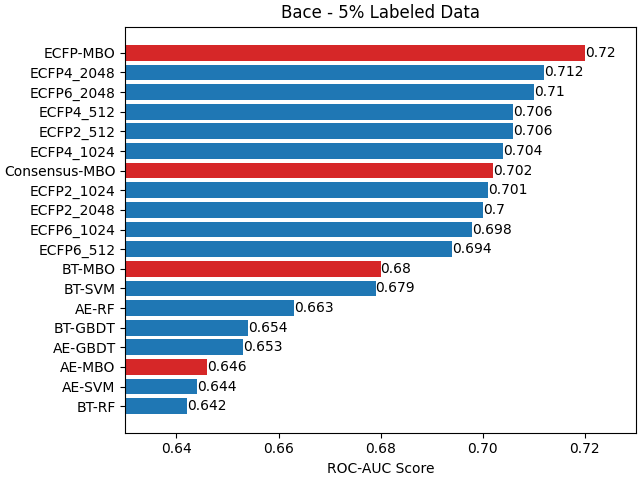}
    \end{minipage}
    \includegraphics[width=0.5\textwidth]{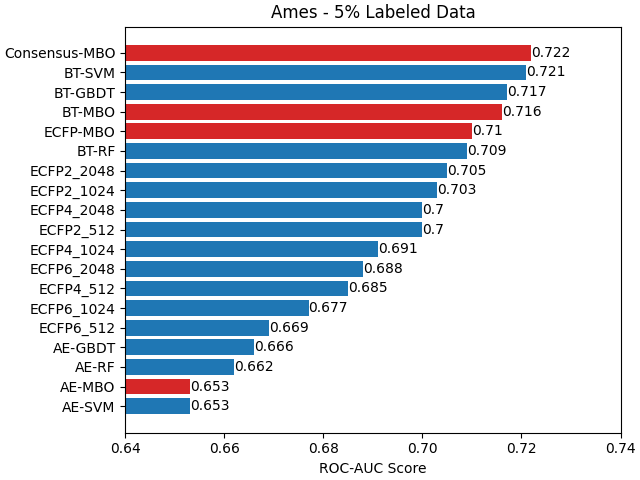}
    \caption{Comparison of our proposed methods (shown in red) with other methods (shown in blue) on the five benchmark data sets for 5\% labeled data.}
    \label{5pfig}
\end{figure}

Table \ref{tab:5 Comparison} and Figure \ref{5pfig} display the results for 5\% labeled data. Again, one of our proposed models outperforms all other comparison methods for all five benchmark data sets. Here, our Consensus-MBO method achieved the best results for both Beet (ROC-AUC = 0.666) and Ames (ROC-AUC = 0.722) data sets, averaging over our ECFP-MBO and BT-MBO methods in both cases. As in the 1\% and 2\% labeled data cases, our BT-MBO method performs significantly better than other comparison methods for the ClinTox (ROC-AUC = 0.774) and BBBP (ROC-AUC = 0.785) data sets. Our proposed ECFP-MBO method again achieved the best performance of all considered methods for the Bace data set (ROC-AUC = 0.72). For the performance of our AE-MBO and BT-MBO models on amounts of labeled data between 5\% and 90\%, see Table \ref{tab:ROC-AUC Scores}.

Chen et al. \cite{hundreds} evaluated their proposed bidirectional encoder transformer models against other models using 90\% labeled data. Table \ref{tab:90 Comparison} displays the performance of the self-supervised learning platform fingerprints (SSLP-FPs) generated from their proposed  transformer model as well as their results using autoencoder-generated fingerprints (Auto-FPs) \cite{auto-fp} and extended-connectivity fingerprints (ECFPs) \cite{circ-fp}. Table \ref{tab:90 Comparison} also displays the performance of our proposed BT-MBO, AE-MBO, and ECFP-MBO models with 90\% labeled data passed to the MBO algorithm. For 90\% labeled data, our proposed BT-MBO method outperforms our AE-MBO method on all data sets, though the AE-MBO method comes closest on the Bace and BBBP data sets. Our ECFP-MBO method outperforms our BT-MBO method on the Bace and Beet data sets and comes very close to the best-performing comparison method on the Beet data set. While the proposed methods perform very well, comparison with other techniques shows that the proposed algorithms are most advantageous when a small amount of data, such as 1\% of the data set, is considered as a labeled set.

\begin{table}[h!]
\vspace{0.5cm}
\centering
\makebox[\textwidth]{
 \begin{tabular}{c|c c c c c}
 \hline
 &\multicolumn{5}{c}{\textbf{ROC-AUC Scores for 90\% Labeled Data}}\\
 \textbf{Model} & \textbf{Ames} & \textbf{Bace} & \textbf{BBBP} & \textbf{Beet} & \textbf{ClinTox}\\
 \hline
 BT-MBO (Proposed) & $0.801 \pm .013$ & $0.839 \pm .02$ & $0.902 \pm .034$ & $0.831 \pm .067$ & $0.948 \pm .048$ \\
 AE-MBO (Proposed) & $0.737 \pm .017$ & $0.818 \pm .037$ & $0.886 \pm .029$ & $0.785 \pm .079$ & $0.807 \pm .085$\\
 ECFP-MBO (Proposed) & $0.787 \pm .019$ & $0.855 \pm .026$ & $0.858 \pm .041$ & $0.842 \pm .051$ & $0.634 \pm .060$ \\
 Consensus-MBO (Proposed) & $0.788 \pm .015$ & $0.833 \pm .030$ & $0.876 \pm .030$ & $0.750 \pm .080$ & $0.819 \pm .015$ \\
 SSLP-FPs(C) \cite{hundreds} & $0.87 \pm .022$ & $0.872 \pm .036$ & $0.949 \pm .016$ & $0.797 \pm .079$ & $\mathbf{0.963 \pm .044}$\\
 SSLP-FPs(CP) \cite{hundreds} & $0.861 \pm .025$ & $0.869 \pm .029$ & $0.953 \pm .009$ & $0.818 \pm .075$ & $0.939 \pm .047$\\
 SSLP-FPs(CPZ) \cite{hundreds} & $0.856 \pm .023$ & $0.868 \pm .033$ & $0.946 \pm .015$ & $0.827 \pm .061$ & $0.941 \pm .035$\\
 Auto-FPs \cite {hundreds, auto-fp} & $\mathbf{0.887 \pm .018}$ & $0.903 \pm .022$ & $\mathbf{0.969 \pm .01}$ & $0.838 \pm .059$ & $0.95 \pm .037$\\
 ECFP2\_512  \cite {hundreds, circ-fp} & $0.867 \pm .016$ & $0.904 \pm .024$ & $0.913 \pm .019$ & $0.83 \pm .06$ & $0.831 \pm .056$\\
 ECFP2\_1024 \cite {hundreds, circ-fp} & $0.868 \pm .016$ & $0.905 \pm .022$ & $0.92 \pm .015$ & $0.839 \pm .058$ & $0.821 \pm .058$\\
 ECFP2\_2048 \cite {hundreds, circ-fp} & $0.869 \pm .02$ & $0.906 \pm .024$ & $0.919 \pm .021$ & $0.833 \pm .063$ & $0.833 \pm .053$\\
 ECFP4\_512  \cite {hundreds, circ-fp} & $0.865 \pm .013$ & $0.896 \pm .023$ & $0.908 \pm .025$ & $0.824 \pm .061$ & $0.801 \pm .049$\\
 ECFP4\_1024 \cite {hundreds, circ-fp} & $0.867 \pm .016$ & $0.901 \pm .024$ & $0.914 \pm .024$ & $0.833 \pm .058$ & $0.782 \pm .052$\\
 ECFP4\_2048 \cite {hundreds, circ-fp} & $0.872 \pm .017$ & $\mathbf{0.907 \pm .023}$ & $0.916 \pm .021$ & $0.828 \pm .066$ & $0.784 \pm .053$\\
 ECFP6\_512  \cite {hundreds, circ-fp} & $0.861 \pm .013$ & $0.889 \pm .026$ & $0.9 \pm .032$ & $\mathbf{0.848 \pm .062}$ & $0.77 \pm .048$\\
 ECFP6\_1024 \cite {hundreds, circ-fp} & $0.862 \pm .013$ & $0.893 \pm .026$ & $0.907 \pm .029$ & $0.833 \pm .059$ & $0.77 \pm .054$\\
 ECFP6\_2048 \cite {hundreds, circ-fp} & $0.864 \pm .014$ & $0.899 \pm .024$ & $0.911 \pm .026$ & $0.828 \pm .063$ & $0.75 \pm .059$\\

 \end{tabular}}
\caption{\label{tab:90 Comparison} Comparison of our models' performance on the five data sets with 90\% labeled data with other models' performance. Our four models are AE-MBO (autoencoder with MBO), BT-MBO (transformer with MBO), ECFP-MBO (extended-connectivity fingerprints with MBO), and Consensus-MBO, a model generating the consensus from our top two scoring methods for a given data set and percent of labeled data (more details in Section \ref{Consensus}). Performance of our models is given as average ROC-AUC score over 10 labeled sets with standard deviation, and Consensus-MBO performance is given as the average over 10 trials with standard deviation.}
\end{table}

\begin{table}[h!]
\vspace{0.5cm}
\centering
\begin{tabular}{c|c c c c c c c c c c}
\hline
  \textbf{Amount} &\multicolumn{10}{c}{\textbf{Data Sets and Models}}\\
\textbf{Labeled} &\multicolumn{2}{c}{\textbf{Ames}} &\multicolumn{2}{c}{\textbf{Bace}} &\multicolumn{2}{c}{\textbf{BBBP}} &\multicolumn{2}{c}{\textbf{Beet}} &\multicolumn{2}{c}{\textbf{ClinTox}}\\
\textbf{Data} & AE & BT & AE & BT & AE & BT & AE & BT & AE & BT\\ [0.5ex]
\hline\hline
90\% & 0.737 & 0.801 & 0.818 & 0.839 & 0.886 & 0.902 & 0.785 & 0.831 & 0.807 & 0.948 \\
80\% & 0.731 & 0.794 & 0.800 & 0.830 & 0.868 & 0.889 & 0.747 & 0.798 & 0.759 & 0.920 \\
75\% & 0.728 & 0.791 & 0.797 & 0.826 & 0.866 & 0.885 & 0.746 & 0.783 & 0.752 & 0.917 \\
70\% & 0.728 & 0.788 & 0.795 & 0.824 & 0.862 & 0.883 & 0.741 & 0.772 & 0.752 & 0.910 \\
60\% & 0.724 & 0.785 & 0.789 & 0.812 & 0.856 & 0.879 & 0.716 & 0.770 & 0.749 & 0.908 \\
50\% & 0.718 & 0.781 & 0.775 & 0.806 & 0.854 & 0.874 & 0.697 & 0.761 & 0.731 & 0.901 \\
40\% & 0.714 & 0.776 & 0.766 & 0.795 & 0.841 & 0.868 & 0.693 & 0.749 & 0.727 & 0.897 \\
30\% & 0.707 & 0.771 & 0.756 & 0.786 & 0.830 & 0.861 & 0.667 & 0.734 & 0.723 & 0.883 \\
25\% & 0.703 & 0.767 & 0.746 & 0.777 & 0.820 & 0.855 & 0.664 & 0.722 & 0.711 & 0.879 \\
20\% & 0.697 & 0.761 & 0.735 & 0.768 & 0.810 & 0.848 & 0.657 & 0.713 & 0.698 & 0.857 \\
10\% & 0.678 & 0.743 & 0.700 & 0.734 & 0.777 & 0.825 & 0.626 & 0.673 & 0.656 & 0.828 \\
5\%  & 0.653 & 0.716 & 0.646 & 0.680 & 0.730 & 0.785 & 0.578 & 0.621 & 0.596 & 0.774 \\
2\%  & 0.619 & 0.677 & 0.589 & 0.618 & 0.685 & 0.736 & 0.548 & 0.576 & 0.561 & 0.704 \\
1\%  & 0.600 & 0.649 & 0.561 & 0.580 & 0.640 & 0.695 & 0.535 & 0.545 & 0.544 & 0.669 \\[1ex]
\hline
\end{tabular}
\caption{\label{tab:ROC-AUC Scores} Performance (given as average ROC-AUC score) of different models on five classification data sets with varying amounts of labeled data. The AE-MBO model's performance is denoted by "AE," and the BT-MBO model's performance is denoted by "BT."}
\end{table}

% \begin{figure}
% \centering
% \begin{subfigure}[t]{\textwidth}
%   \centering
%   \includegraphics[width=0.9\linewidth]{bbbp_plots.png}
%   \label{fig:bbbp}
% \end{subfigure}%
% \vspace{8pt}
% \begin{subfigure}[t]{\textwidth}
%   \centering
%   \includegraphics[width=0.9\linewidth]{clintox_plots.png}
%   \label{fig:clintox}
% \end{subfigure}
% \begin{subfigure}[t]{\textwidth}
%   \centering
%   \includegraphics[width=0.9\linewidth]{beet_plots.png}
%   \label{fig:beet}
% \end{subfigure}
% \begin{subfigure}[t]{\textwidth}
%   \centering
%   \includegraphics[width=0.9\linewidth]{bace_plots.png}
%   \label{fig:bace}
% \end{subfigure}
% \begin{subfigure}[t]{\textwidth}
%   \centering
%   \includegraphics[width=0.9\linewidth]{ames_plots.png}
%   \label{fig:ames}
% \end{subfigure}
% \caption{Comparison of our methods with other methods (self-supervised learning fingerprints and autoencoder fingerprints). The proposed methods are shown in red, and other methods are shown in blue.}
% \label{fig:results}
% \end{figure}

% \begin{figure}
%     \centering
%     \begin{subfigure}[t]{0.45\textwidth}
%         \includegraphics[width = \linewidth]{clintox1_plot.png}
%     \end{subfigure}
%     \hfill
%     \begin{subfigure}[t]{0.45\textwidth}
%         \includegraphics[width = \linewidth]{bbbp1_plot.png}
%     \end{subfigure}
%     \caption{Caption}
%     \label{fig:clintox1}
% \end{figure}

To further evaluate our proposed models, we utilize the R-S scores and R-S index discussed in Section \ref{evaluation}. Recall that the R-S index for a given data set and model (i.e., a given featurization of a data set) essentially measures the closeness of that data set's residue index and similarity index--if the residue and similarity indices are close together in value, the R-S index will be close to $1$, while dissimilar residue and similarity indices produce an R-S index closer to $0$. Further recall that residue scores measure how far a class is from other classes, and similarity scores measure how closely clustered data are in a class. 

\begin{figure}
    \centering
    \includegraphics[width=0.5\textwidth]{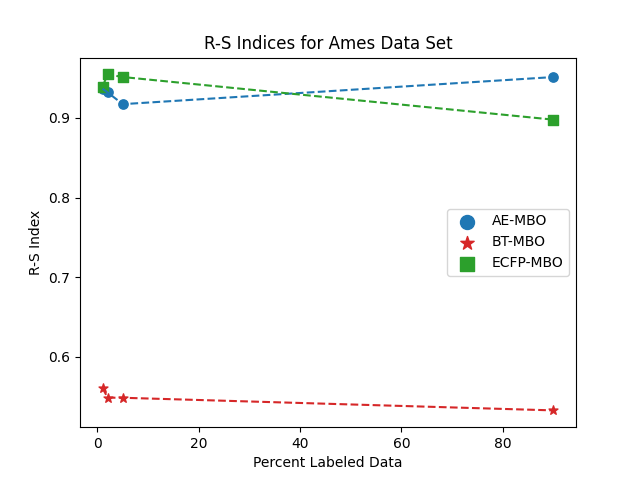}
    \includegraphics[width=0.6\textwidth]{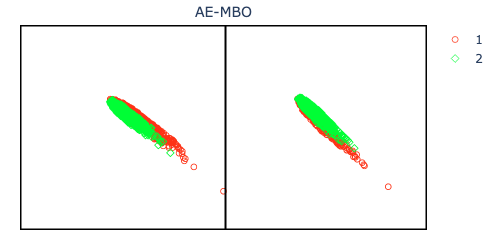}
    \includegraphics[width=0.6\textwidth]{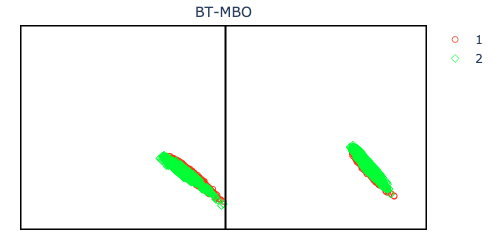}
    \includegraphics[width=0.6\textwidth]{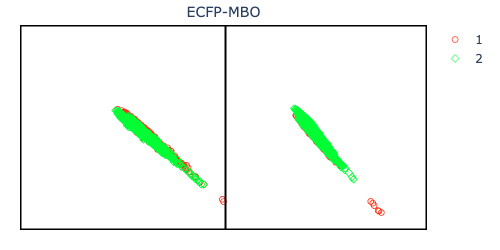}
    \caption{R-S index and R-S score plots for the Ames data set. The R-S index plot displays R-S indices for our AE-MBO, BT-MBO, and ECFP-MBO models for 1\%, 2\%, 5\%, and 90\% labeled data. The three R-S score plots display R-S scores for a random split of the Ames data set with 5\% labeled data for AE-MBO, BT-MBO, and ECFP-MBO, from top to bottom. Each panel plots data points from one class, and the points are colored by their predicted class. The $x$ and $y$ axes represent the residue and similarity scores, respectively.}
    \label{fig:ames_rs_plots}
\end{figure}

\begin{figure}
    \centering
    \includegraphics[width=0.5\textwidth]{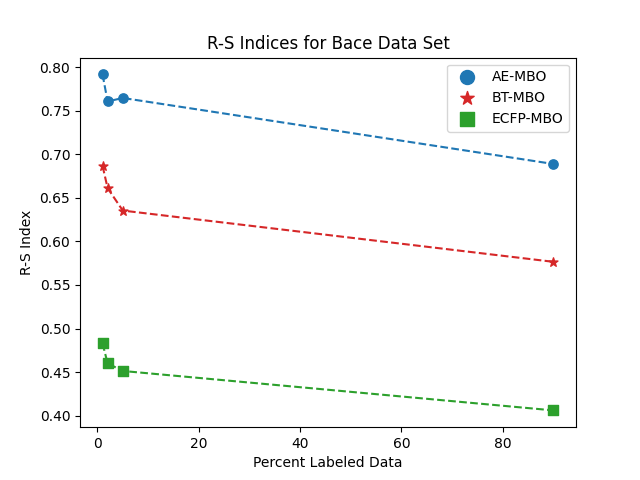}
    \includegraphics[width=0.6\textwidth]{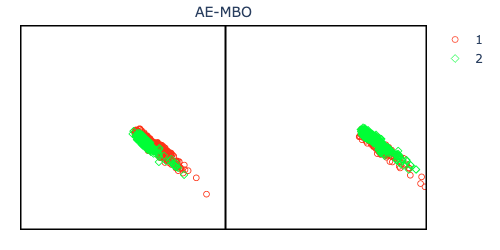}
    \includegraphics[width=0.6\textwidth]{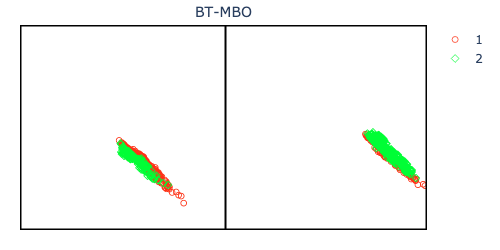}
    \includegraphics[width=0.6\textwidth]{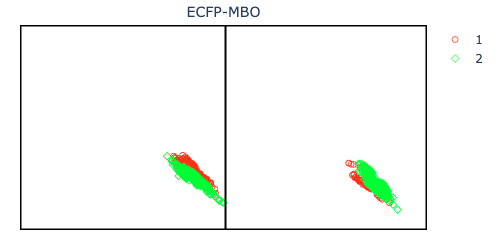}
    \caption{R-S index and R-S score plots for the Bace data set. The R-S index plot displays R-S indices for our AE-MBO, BT-MBO, and ECFP-MBO models for 1\%, 2\%, 5\%, and 90\% labeled data. The three R-S score plots display R-S scores for a random split of the Bace data set with 5\% labeled data for AE-MBO, BT-MBO, and ECFP-MBO, from top to bottom. Each panel plots data points from one class, and the points are colored by their predicted class. The $x$ and $y$ axes represent the residue and similarity scores, respectively.}
    \label{fig:bace_rs_plots}
\end{figure}

\begin{figure}
    \centering
    \includegraphics[width=0.5\textwidth]{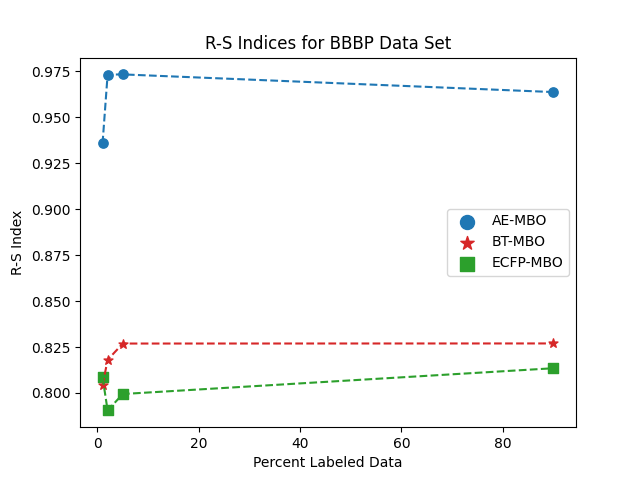}
    \includegraphics[width=0.6\textwidth]{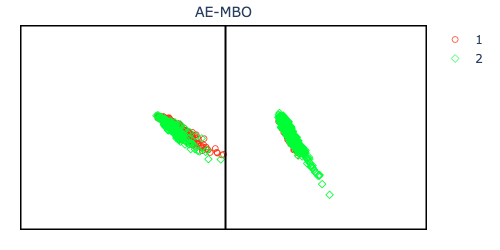}
    \includegraphics[width=0.6\textwidth]{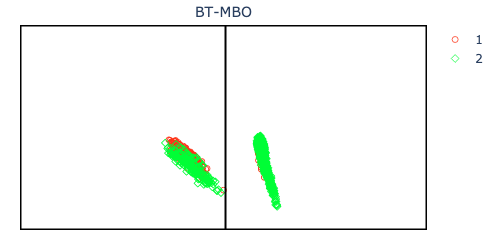}
    \includegraphics[width=0.6\textwidth]{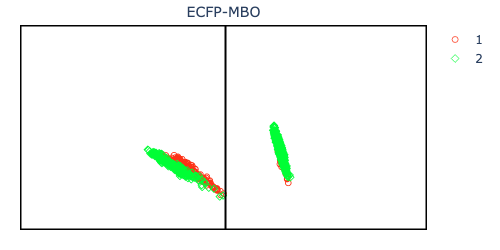}
    \caption{R-S index and R-S score plots for the BBBP data set. The R-S index plot displays R-S indices for our AE-MBO, BT-MBO, and ECFP-MBO models for 1\%, 2\%, 5\%, and 90\% labeled data. The three R-S score plots display R-S scores for a random split of the BBBP data set with 5\% labeled data for AE-MBO, BT-MBO, and ECFP-MBO, from top to bottom. Each panel plots data points from one class, and the points are colored by their predicted class. The $x$ and $y$ axes represent the residue and similarity scores, respectively.}
    \label{fig:bbbp_rs_plots}
\end{figure}

\begin{figure}
    \centering
    \includegraphics[width=0.5\textwidth]{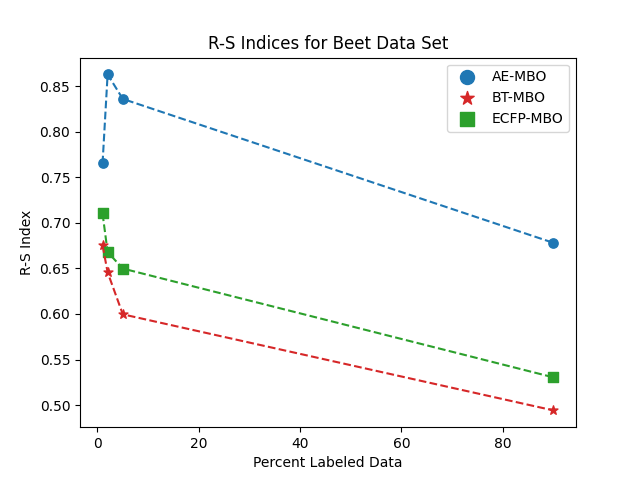}
    \includegraphics[width=0.6\textwidth]{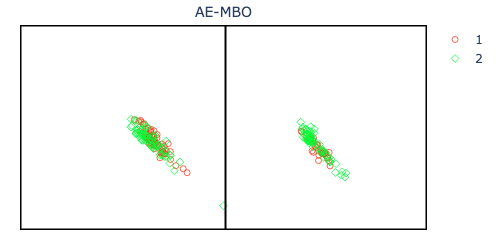}
    \includegraphics[width=0.6\textwidth]{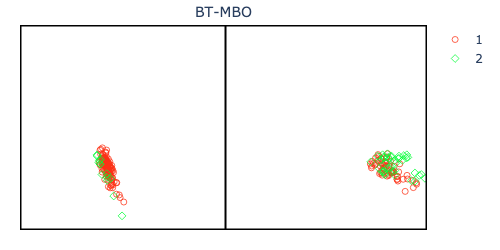}
    \includegraphics[width=0.6\textwidth]{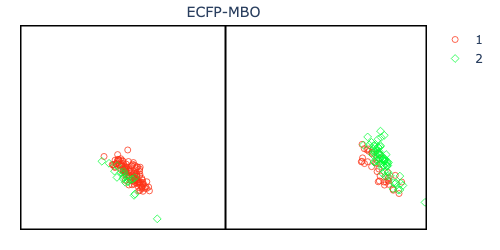}
    \caption{R-S index and R-S score plots for the Beet data set. The R-S index plot displays R-S indices for our AE-MBO, BT-MBO, and ECFP-MBO models for 1\%, 2\%, 5\%, and 90\% labeled data. The three R-S score plots display R-S scores for a random split of the Beet data set with 5\% labeled data for AE-MBO, BT-MBO, and ECFP-MBO, from top to bottom. Each panel plots data points from one class, and the points are colored by their predicted class. The $x$ and $y$ axes represent the residue and similarity scores, respectively.}
    \label{fig:beet_rs_plots}
\end{figure}

\begin{figure}
    \centering
    \includegraphics[width=0.5\textwidth]{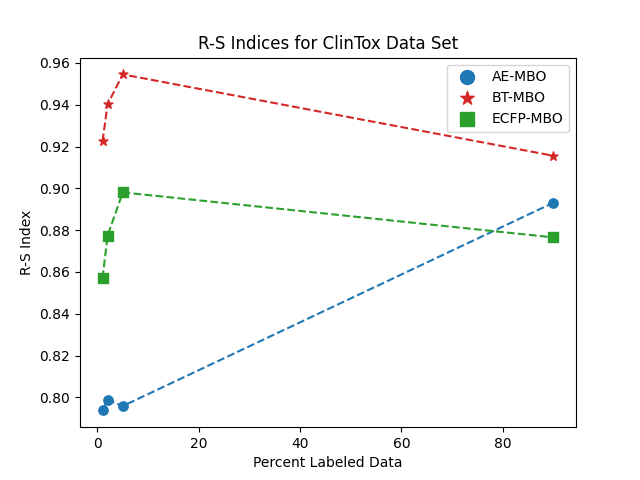}
    \includegraphics[width=0.6\textwidth]{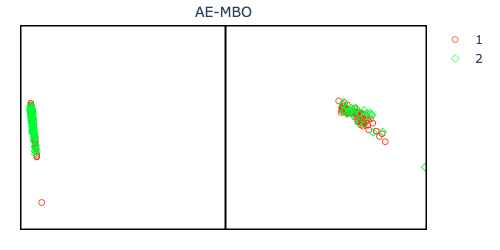}
    \includegraphics[width=0.6\textwidth]{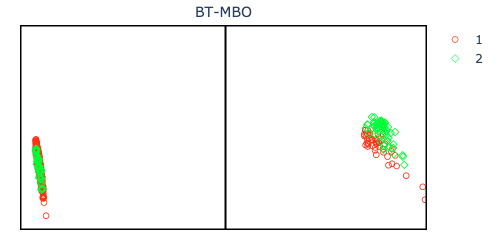}
    \includegraphics[width=0.6\textwidth]{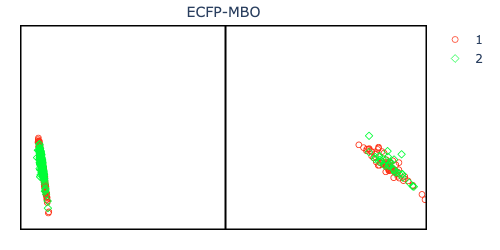}
    \caption{R-S index and R-S score plots for the ClinTox data set. The R-S index plot displays R-S indices for our AE-MBO, BT-MBO, and ECFP-MBO models for 1\%, 2\%, 5\%, and 90\% labeled data. The three R-S score plots display R-S scores for a random split of the ClinTox data set with 5\% labeled data for AE-MBO, BT-MBO, and ECFP-MBO, from top to bottom. Each panel plots data points from one class, and the points are colored by their predicted class. The $x$ and $y$ axes represent the residue and similarity scores, respectively.}
    \label{fig:clintox_rs_plots}
\end{figure}

Figures \ref{fig:ames_rs_plots}, \ref{fig:bace_rs_plots}, \ref{fig:bbbp_rs_plots}, \ref{fig:beet_rs_plots}, and \ref{fig:clintox_rs_plots} display R-S index and R-S score plots for our AE-MBO, BT-MBO, and ECFP-MBO models on the five benchmark data sets used in this work. For a given data set and model, the R-S index was calculated for each training/testing split of the data, and then the average R-S index was calculated over 50 random splits for 1\%, 2\%, and 5\% labeled data and over 10 random splits for 90\% labeled data. R-S scores were calculated on a random training/testing split with 5\% labeled data. The panels in the R-S score plots display data points in the two classes, and each point is assigned a color based on its predicted class from the model. The $x$ and $y$ axes in the R-S score plots represent the residue and similarity scores, respectively.

For the Ames data set, the best-performing model (in terms of ROC-AUC score) for 1\% labeled data was the BT-MBO model, with the ECFP-MBO model following close behind and the AE-MBO model performing significantly worse. The same trend can be seen for 2\%, 5\%, and 90\% labeled data. In Figure \ref{fig:ames_rs_plots}, we see that the BT-MBO method has markedly lower R-S indices for the Ames data set across all percentages of labeled data. Considering the R-S score plots, the BT-MBO method displays a higher degree of dissimilarity between its residue and similarity scores, which aligns with the low R-S index values. Furthermore, the BT-MBO model has higher residue scores than the other two models, indicating that the two classes are well separated from each other. This may contribute to the BT-MBO model's superior performance.

Our proposed ECFP-MBO method consistently had higher ROC-AUC scores than the AE-MBO and BT-MBO methods on the Bace data set for 1\%, 2\%, 5\%, and 90\% labeled data. Figure \ref{fig:bace_rs_plots} shows that the ECFP-MBO method has the lowest R-S indices for all amounts of labeled data. The points in the R-S score plot for the ECFP-MBO model are concentrated in the lower right corner for both classes, reflecting that the Bace data points, on average, have high residue scores and low similarity scores. The high residue scores for this model correlate with its improved performance compared to the other models.

Examining Figure \ref{fig:bbbp_rs_plots} for the BBBP data set, we begin to see different R-S score distributions for the two classes. The three models all generally performed well in making predictions the BBBP data set. In fact, for a given amount of labeled data, one of our models almost always achieved the highest ROC-AUC scores on the BBBP data set compared to the other four data sets (with the only exception being the ClinTox data set for 90\% labeled data). Compared to the previous two data sets, the BBBP data set shows higher R-S indices for all three models and all percentages of labeled data. We do see less obvious dissimilarity between the residue and similarity scores for the three models in Figure \ref{fig:bbbp_rs_plots}, which corresponds to higher R-S index values. We also observe that data points in the first class tend to have high residue scores, indicating that points in this class are far from points in the second class. However, points in the second class have comparatively lower residue scores with varying similarity scores. 

For the Beet data set, our ECFP-MBO method achieved higher ROC-AUC scores than the AE-MBO and BT-MBO methods for all tasks, and the AE-MBO model consistently performed the worst. Comparing the R-S indices shown in Figure \ref{fig:beet_rs_plots}, the AE-MBO method has markedly higher R-S indices than the other two methods for all percentages of labeled data. This reflects the closeness of the residue and similarity scores for the AE-MBO method; the points on the R-S score plot for the AE-MBO method are much more centrally located than those in the other two plots. We can further observe that the AE-MBO method seems to have more data point misclassifications than the other two methods. Specifically, the AE-MBO model accurately classified most points in the second class but misclassified a large portion of points in the first class. Furthermore, the ECFP-MBO method appears to have fewer misclassifications across the two classes than the other two models.

The ClinTox data set displays a significant difference in R-S score distribution for its two classes, as seen in Figure \ref{fig:clintox_rs_plots}. Recall that, for all percentages of labeled data, the BT-MBO model performed significantly better with respect to ROC-AUC score than all other models on the ClinTox data set. We observe that the BT-MBO model also has very high R-S index values for all amounts of labeled data. Although data points in the first class tend to have low residue scores for all three featurizations, data points in the second class tend to have high residue scores. A high R-S index corresponds to residue and similarity indices that are close together. Because the residue index and similarity index are properties of the entire data set and involve averaging across classes, low residue scores from one class can be offset by high residue scores from another class. In the case of the ClinTox data set as evidenced by the R-S score plots in Figure \ref{fig:clintox_rs_plots}, the residue index and similarity index would then end up being similar in magnitude, resulting in a high R-S index value. Furthermore, the high residue scores for data points in the second class may contribute to all three models' superior performance on this data set.  

% \begin{figure}
%     \centering
%     \begin{minipage}{0.5\textwidth}
%         \centering
%         \includegraphics[width=\textwidth]{rs_index_plot_ames.png}
%     \end{minipage}\hfill
%     \begin{minipage}{0.5\textwidth}
%         \centering
%         \includegraphics[width=\textwidth]{rs_index_plot_bace.png}
%     \end{minipage}
%     \begin{minipage}{0.5\textwidth}
%         \centering
%         \includegraphics[width=\textwidth]{rs_index_plot_bbbp.png}
%     \end{minipage}\hfill
%     \begin{minipage}{0.5\textwidth}
%         \centering
%         \includegraphics[width=\textwidth]{rs_index_plot_beet.png}
%     \end{minipage}
%     \includegraphics[width=0.5\textwidth]{rs_index_plot_clintox.png}
%     \caption{R-S indices for all five benchmark data sets and our AE-MBO, BT-MBO, and ECFP-MBO models for 1\%, 2\%, 5\%, and 90\% labeled data. For a given data set and model, the R-S index was calculated for each training/testing split of the data, and then the average R-S index was calculated over 50 random splits for 1\%, 2\%, and 5\% labeled data and over 10 random splits for 90\% labeled data.}
%     \label{rs-index-fig}
% \end{figure}

\section{Conclusion}

Given the difficulty of obtaining labeled molecular data, particularly very large amounts of labeled molecular data, the ability to accurately predict molecular properties with limited labeled data can be crucial. This paper presents three novel techniques, namely the BT-MBO, AE-MBO, and ECFP-MBO algorithms in addition to a consensus model (Consensus-MBO), which perform very well even with small amounts of labeled data. In particular, our proposed methods achieve better or comparable ROC-AUC scores compared to state-of-the-art methods when as little as 1\% of the data set is designated as a labeled set. Notably, for the ClinTox and BBBP data sets, all four of our proposed methods (including the Consensus-MBO method) consistently rank among the top four or five of all comparison methods for 1\%, 2\%, and 5\% labeled data. Moreover, the proposed BT-MBO, ECFP-MBO, and Consensus-MBO methods achieved the best performance for a benchmark data set in at least one of the low labeled data experiments. The analysis of residue and similarity (R-S) scores and R-S indices offers an explanation of the observed performance. Overall, the three new MBO-assisted models in this work, together with the proposed Consensus-MBO technique, can serve as powerful predictive tools when labeled data is scarce.

\section*{Data Availability}
The five classification data sets (Ames, Bace, BBBP, Beet, ClinTox) used in this work are available in SMILES string format at \url{https://weilab.math.msu.edu/DataLibrary/2D/}.

\section*{Code Availability}
The source code used to generate features for the data sets can be found on GitHub at \url{https://github.com/WeilabMSU/antoencoder-v01} for the autoencoder (AE-MBO) model and \url{https://github.com/WeilabMSU/PretrainModels} for the transformer (BT-MBO) model.

\section*{Supporting Information}
The optimal hyperparameters of the proposed methods are reported in the Supporting Information document.

\section*{Conflict of Interest}
The authors declare no conflict of interest.

\section*{Acknowledgements}
This work is supported in part by NSF grant DMS-2052983.
The work of   Guo-Wei Wei was supported in part by NIH grants  R01GM126189 and  R01AI164266, NSF grants DMS-2052983,  DMS-1761320, and IIS-1900473,  NASA grant 80NSSC21M0023,  Michigan Economic Development Corporation, MSU Foundation,  Bristol-Myers Squibb 65109, and Pfizer.

\newpage

\bibliographystyle{plain}
%\bibliographystyle{unsrt}
%\bibliography{bib_draft.bib}
\bibliography{bib_draft}
% can add multiple bib files in separate lines

\end{document}